\documentclass[accepted]{uai2023}

\usepackage{xr-hyper}
\usepackage[table]{xcolor}
\usepackage[american]{babel}
\usepackage{natbib} 
\usepackage{mathtools} 
\usepackage{booktabs} 
\usepackage{tikz} 
\usepackage{amsthm}

\newcommand*\samethanks[1][\value{footnote}]{\footnotemark[#1]}

\title{Functional Causal Bayesian Optimization}

\author[1,\thanks{Equal contribution.}]{Limor Gultchin}
\author[2,\samethanks]{Virginia Aglietti}
\author[2]{Alexis Bellot}
\author[2]{Silvia Chiappa}
\affil[1]{University of Oxford, The Alan Turing Institute, Work done at DeepMind, London, UK}
\affil[2]{DeepMind, London, UK}

\usepackage[capitalize,noabbrev]{cleveref}
\usepackage{graphicx}

\theoremstyle{plain} 
\newcommand{\thistheoremname}{}
\newtheorem*{genericthm*}{\thistheoremname}
\newenvironment{namedthm*}[1]
  {\renewcommand{\thistheoremname}{#1}%
  \begin{genericthm*}}
  {\end{genericthm*}}

\newtheorem{theorem}{Theorem}[section]
\newtheorem{proposition}[theorem]{Proposition}

\theoremstyle{definition}
\newtheorem{definition}[theorem]{Definition}

\theoremstyle{remark}

\usepackage{amsfonts,dsfont}
\usepackage[]{xcolor}
\definecolor{darkPurple}{HTML}{423C70}
\definecolor{darkGreen}{HTML}{00544B}
\definecolor{cboGreen}{HTML}{045D04}
\definecolor{lightGreen}{HTML}{ADDC37}
\definecolor{brightBlue}{HTML}{47A6FB}
\definecolor{cocaBOPurple}{HTML}{95A5A6}
\definecolor{mcbosoftpurple}{HTML}{BB8FCE}

\usepackage{wrapfig}
\usepackage{tikz}
\usepackage[colorinlistoftodos, textwidth=26mm, shadow, color=blue!30!white, textsize=tiny]{todonotes}
\usepackage{cancel} 
\usepackage{dashrule}
\usetikzlibrary{arrows,shapes,backgrounds,through,shadows}
\usetikzlibrary{decorations.pathmorphing,calc}

\usepackage{mathrsfs}
\usepackage{bbm}
\usepackage{bm}

\tikzset{
  dot node/.style={
    shape=circle,
    draw,
    inner sep=+0pt,
    minimum size=+4.mm
  },
  dotdot node/.style 2 args={
    dot node,
    label={[shape=circle,fill=gray,outer sep=+0pt,inner sep=+0pt,minimum size=+2.mm]center:}
  },
  arc style/.style={
    |<->|,
    shorten >=+-.5\pgflinewidth,
    shorten <=+-.5\pgflinewidth,
  }
}

\usepackage{array,multirow,graphicx}
\usepackage{float}
\usepackage[algo2e]{algorithm2e} 
\usepackage{algorithm, algorithmic}
\usepackage{enumitem}
\usepackage{tikz-network}
\usepackage{amssymb}
\usetikzlibrary{automata,positioning}
\tikzstyle{dot}=[circle,fill,inner sep=2.5pt]  
\tikzstyle{dgraph}=[->, line width=1.5pt]
\newcommand{\arr}{-{Triangle[length=2mm, width=2mm]}}
\newcommand{\indep}{\rotatebox[origin=c]{90}{$\models$}}
\renewcommand{\eqref}[1]{Eq. (\ref{#1})}
\newcommand{\figref}[1]{Fig. \ref{#1}}
\newcommand{\secref}[1]{Section \ref{#1}}

\DeclareMathOperator*{\argmin}{arg\,min}
\DeclareMathOperator*{\argmax}{arg\,max}
\DeclareBoldMathCommand{\A}{A}
\DeclareBoldMathCommand{\X}{X}
\DeclareBoldMathCommand{\x}{x}
\DeclareBoldMathCommand{\Z}{Z}
\DeclareBoldMathCommand{\z}{z}
\DeclareBoldMathCommand{\I}{I}
\DeclareBoldMathCommand{\U}{U}
\DeclareBoldMathCommand{\V}{V}
\DeclareBoldMathCommand{\F}{F}
\DeclareBoldMathCommand{\C}{C}
\DeclareBoldMathCommand{\c}{c}
\newcommand{\datai}{\mathcal{D}^I}
\newcommand{\acronospace}[1]{\textsc{#1}}
\newcommand{\gpstext}{\acronospace{gp}s}
\newcommand{\graph}{\mathcal{G}}

\newcommand{\boldomega}{\bm{\omega}}
\newcommand{\range}{\mathcal{R}}
\newcommand{\cond}{\,|\,}

\newcommand{\acro}[1]{\textsc{#1}\xspace}
\newcommand{\ie}{i.e.\xspace}
\newcommand{\eg}{e.g.\xspace}
\newcommand{\doi}{\text{do}}
\newcommand{\scm}{\acro{scm}}
\newcommand{\scms}{\acronospace{scm}s}

\newcommand{\gptext}{\acro{gp}}
\newcommand{\rbf}{\acro{rbf}}
\newcommand{\psa}{\acro{psa}}
\newcommand{\bmi}{\acro{bmi}}
\newcommand{\bmr}{\acro{bmr}}
\newcommand{\rkhs}{\acro{rkhs}}
\newcommand{\mps}{\acro{mps}}
\newcommand{\dmp}{\acro{dmp}}

\newcommand{\ci}{\acro{ci}}
\newcommand{\Age}{\text{Age}}

\newcommand{\Aspirin}{\text{Aspirin}}
\newcommand{\Statin}{\text{Statin}}
\newcommand{\Height}{\text{Height}}
\newcommand{\health}{\acro{health}}
\newcommand{\echain}{\acro{chain}}
\newcommand{\st}{*}

\newcommand{\calF}{\mathcal{F}}
\newcommand{\Sset}{\mathcal{S}}
\newcommand{\piS}{\mathbf{\pi}_{\Sset}}

\newcommand{\PiS}{\Pi_{\Sset}}

\newcommand{\cvalue}{\mathbf{c}}

\newcommand{\mismps}{{\mathbb{M}}_{\Sigma}}

\newcommand{\boldalpha}{\boldsymbol{\alpha}}
\newcommand{\boldbeta}{\boldsymbol{\beta}}

\newcommand{\cgo}{\acro{cgo}}
\newcommand{\fcgo}{f\acro{cgo}}
\newcommand{\fgo}{\acro{fgo}}
\newcommand{\cbo}{\acro{cbo}}
\newcommand{\fcbo}{f\acro{cbo}}
\newcommand{\cocabo}{\acronospace{c}o\acronospace{c}a-\bo}
\newcommand{\fbo}{\acro{bfo}}
\newcommand{\bo}{\acro{bo}}
\newcommand{\mcbo}{\acro{mcbo}}
\newcommand{\fei}{f\acro{ei}}

\newcommand{\hardsubscript}{\text{hard}}
\newcommand{\softsubscript}{\text{func}}
\newcommand{\functsubscript}{\text{func}}

\newcommand{\cost}{\texttt{Co}}

\newcommand{\langerangle}[2]{\langle #1, #2 \rangle}

\newcommand{\pa}{\text{pa}}
\newcommand{\opa}{\text{pa}}
\newcommand{\an}{\text{an}}
\newcommand{\de}{\text{de}}
\newcommand{\spo}{\text{sp}}
\def\*#1{\mathbf{#1}}
\def\1#1{\mathcal{#1}}
\def\2#1{\mathscr{#1}}
\def\3#1{\mathbb{#1}}
\newcommand{\kappapar}{\xi}

\usepackage{multicol,multirow} \usepackage{hhline} 
\newcommand{\nrmpsreduce}{\texttt{NRMPSReduce}}
\usepackage{subfig}
\usepackage{graphicx}
\usepackage[most]{tcolorbox}

\begin{document}
\maketitle

\begin{abstract}
We propose \emph{functional causal Bayesian optimization} (\fcbo), a method for finding interventions that optimize a target variable in a known causal graph. \fcbo extends the \cbo family of methods to enable functional interventions, which set a variable to be a deterministic function of other variables in the graph. \fcbo models the unknown objectives with Gaussian processes whose inputs are defined in a reproducing kernel Hilbert space, thus allowing to compute distances among vector-valued functions. In turn, this enables to sequentially select functions to explore by maximizing an expected improvement acquisition functional while keeping the typical computational tractability of standard \bo settings. We introduce graphical criteria that establish when considering functional interventions allows attaining better target effects, and conditions under which selected interventions are also optimal for conditional target effects. We demonstrate the benefits of the method in a synthetic and in a real-world causal graph.
\end{abstract}

\section{Introduction}\label{sec:introduction}
Finding interventions in a system that optimize a target variable is key to many scientific disciplines, including medicine, biology, and social sciences.
Causal graphs \citep{pearl2000causality, kollerl2009probabilistic}, in which an intervention on a variable is represented as modifying the casual influence from its incoming edges, offer a powerful tool for dealing with the effects of interventions, and are therefore increasingly integrated into approaches to learning optimal policies such as bandits \citep{lattimore2016causal, lee2018structural, lee2019structural, lu2020regret, nair2021budgeted, de2022causal}, reinforcement learning \citep{lu2018deconfounding, zhang2019near, zhang2020designing, gasse2021causal, zhang2022online}, and Bayesian optimization \citep{aglietti2020causal, aglietti2021dynamic, sussex2022model}. 

Most works in causal Bayesian optimization (\cbo) have focused on the \emph{hard intervention} $\doi(X=x)$, which consists in setting variable $X$ to a constant value $x$. However, in many practical scenarios the investigator may be able to implement policies that also contain other types of interventions.
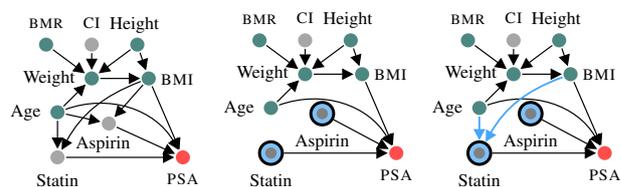
\begin{figure}[t]
\centering
\scalebox{0.75}{\begin{tikzpicture}[dgraph]
\node[dot] (cal) [fill=gray!70,label=north:\ci] at (0, 2) {};
\node[dot] (bmr) [fill=darkGreen!70,label=north:\small{\acro{bmr}}] at (-0.8, 2) {};
\node[dot] (age) [fill=darkGreen!70,label=left:\small{Age}] at (-0.6,0.8) {};
\node[dot] (weight) [fill=darkGreen!70,label=left:\small{Weight}] at (0, 1.4) {};
\node[dot] (height) [fill=darkGreen!70,label={[xshift=-0.05cm, yshift=-0.08cm]\small{Height}}] at (0.8, 2) {};
\node[dot] (BMI)[fill=darkGreen!70, label={[xshift=0.5cm, yshift=-0.4cm]:\bmi}] at (1,1.4) {};
\node[dot] (Aspirin) [fill=gray!70,label={[xshift=-0.1cm, yshift=-0.8cm]\small{Aspirin}}] at (0.3,0.6) {};
\node[dot] (Statin) [fill=gray!70,label=south:\small{Statin}] at (-0.6, 0) {};
\node[dot] (PSA) [fill=red!70,label=south:\psa] at (1.6,0) {};
\draw[line width=0.6pt, \arr](age)--(weight);
\draw[line width=0.6pt, \arr](bmr)--(weight);
\draw[line width=0.6pt, \arr](cal)--(weight);
\draw[line width=0.6pt, \arr](height)--(weight);
\draw[line width=0.6pt, \arr](height)--(BMI);
\draw[line width=0.6pt, \arr](weight)--(BMI);
\draw[line width=0.6pt, \arr](BMI)--(PSA);
\draw[line width=0.6pt, \arr](BMI)--(Aspirin);
\draw[line width=0.6pt, \arr](age)--(Aspirin);
\draw[line width=0.6pt, \arr](age)--(Statin);
\draw[line width=0.6pt, \arr](Aspirin)--(PSA);
\draw[line width=0.6pt, \arr](Statin)--(PSA);
\draw[line width=0.6pt, \arr](BMI)to [bend right=+20](Statin);
\draw[line width=0.6pt, \arr](age)to [bend left=+40](PSA);
\end{tikzpicture}}
\scalebox{0.75}{
\begin{tikzpicture}[dgraph]
\node[dot] (cal) [fill=gray!70,label=north:\ci] at (0, 2) {};
\node[dot] (bmr) [fill=darkGreen!70,label=north:\small{\acro{bmr}}] at (-0.8, 2) {};
\node[dot] (age) [fill=darkGreen!70,label=left:\small{Age}] at (-0.6,0.8) {};
\node[dot] (weight) [fill=darkGreen!70,label=left:\small{Weight}] at (0, 1.4) {};
\node[dot] (height) [fill=darkGreen!70,label={[xshift=-0.05cm, yshift=-0.08cm]\small{Height}}] at (0.8, 2) {};
\node[dot] (BMI)[fill=darkGreen!70, label={[xshift=0.5cm, yshift=-0.4cm]:\bmi}] at (1,1.4) {};
\node[dotdot node] (Aspirin) [fill=brightBlue!70, label=south:\small{Aspirin}] at (0.3, 0.7) {};
\node[dotdot node] (Statin) [fill=brightBlue!70, label=south:\small{Statin}] at (-0.6, 0) {};
\node[dot] (PSA) [fill=red!70,label=south:\psa] at (1.6,0) {};
\draw[line width=0.6pt, \arr](age)--(weight);
\draw[line width=0.6pt, \arr](bmr)--(weight);
\draw[line width=0.6pt, \arr](cal)--(weight);
\draw[line width=0.6pt, \arr](height)--(weight);
\draw[line width=0.6pt, \arr](height)--(BMI);
\draw[line width=0.6pt, \arr](weight)--(BMI);
\draw[line width=0.6pt, \arr](BMI)--(PSA);
\draw[line width=0.6pt, \arr](Aspirin)--(PSA);
\draw[line width=0.6pt, \arr](Statin)--(PSA);
\draw[line width=0.6pt, \arr](age)to [bend left=+40](PSA);
\end{tikzpicture}}
\scalebox{0.75}{\begin{tikzpicture}[dgraph]
\node[dot] (cal) [fill=gray!70,label=north:\ci] at (0, 2) {};
\node[dot] (bmr) [fill=darkGreen!70,label=north:\small{\acro{bmr}}] at (-0.8, 2) {};
\node[dot] (age) [fill=darkGreen!70,label=left:\small{Age}] at (-0.6,0.8) {};
\node[dot] (weight) [fill=darkGreen!70,label=left:\small{Weight}] at (0, 1.4) {};
\node[dot] (height) [fill=darkGreen!70,label={[xshift=-0.05cm, yshift=-0.08cm]\small{Height}}] at (0.8, 2) {};
\node[dot] (BMI)[fill=darkGreen!70, label={[xshift=0.5cm, yshift=-0.4cm]:\bmi}] at (1,1.4) {};
\node[dotdot node] (Aspirin) [fill=brightBlue!70, label=south:\small{Aspirin}] at (0.3, 0.7) {};
\node[dotdot node] (Statin) [fill=brightBlue!70, label=south:\small{Statin}] at (-0.6, 0) {};
\node[dot] (PSA) [fill=red!70,label=south:\psa] at (1.6,0) {};
\draw[line width=0.6pt, \arr](age)--(weight);
\draw[line width=0.6pt, \arr](bmr)--(weight);
\draw[line width=0.6pt, \arr](cal)--(weight);
\draw[line width=0.6pt, \arr](height)--(weight);
\draw[line width=0.6pt, \arr](height)--(BMI);
\draw[line width=0.6pt, \arr](weight)--(BMI);
\draw[line width=0.6pt, \arr](BMI)--(PSA);
\draw[line width=0.6pt, \arr](Aspirin)--(PSA);
\draw[line width=0.6pt, \arr](Statin)--(PSA);
\draw[line width=0.6pt, \arr](age)to [bend left=+40](PSA);
\draw[line width=0.9pt, \arr, brightBlue](age) -- (Statin);
\draw[line width=0.9pt, \arr, brightBlue](BMI)to [bend right=+20](Statin);
\end{tikzpicture}}
\caption{\begin{small} 
\emph{Left:} Graph representing causal relationships between prostate specific antigen (\psa) and other variables. Red, grey and green nodes indicate target, intervenable, non-intervenable variables respectively.
\emph{Middle:} Modified graph describing a policy made of hard interventions on Aspirin and Statin.   
\emph{Right:}  Modified graph describing a policy with an intervention on Statin that retains dependence on Age and \bmi. 
\end{small} 
}
\label{fig:causalgraphs1}
\end{figure}
Consider, for example, the graph in \figref{fig:causalgraphs1}(left) representing causal relationships between prostate specific antigen (\psa) and other variables. An investigator wishing to find a policy for prescribing Aspirin and Statin dosages, as well as Calories Intake (\ci), that minimizes \psa might be able to consider, in addition to policies made of only hard interventions (as the one represented in \figref{fig:causalgraphs1}(middle)), also policies where e.g. Statin dosage retains a dependence on Age and \bmi (as the one represented in \figref{fig:causalgraphs1}(right)).

Contextual interventions are achieved in \citet{arsenyan2023contextual} and in \citet{sussex2022model} by searching for different hard interventions in separate sub-groups defined by some contexts and by inducing changes in the parametrization of a node's conditional distribution via action variables, respectively. However, the first approach learns an implicit mapping between contexts and intervention values, and requires extrapolating to unseen or rarely explored areas of the context space; while the second approach can only induce some modifications of the parametrization and does not allow choice of context. 

In this work, we introduce an extension of the \cbo family of methods that considers a more flexible and general type of contextual intervention, consisting in making variable $X$ a \emph{deterministic function} of other nodes in the graph. Such a \emph{functional intervention} is implemented via new techniques for computing distances among functions of different variables. 
Our contributions can be summarized as follows: 
\begin{itemize}[leftmargin=0.35cm]
\item We formalize the problem of finding policies made of hard and functional interventions optimizing the expectation of a target variable as the \emph{functional causal global optimization} (\fcgo) problem. 
\item  We introduce two graphical criteria that establish when functional interventions could be necessary to solve the \fcgo problem and when policies made of only hard interventions are sufficient, respectively.
\item We introduce conditions in which a policy solving the \fcgo problem also optimizes conditional expectations of the target variable. 
\item We propose \emph{functional causal Bayesian optimization} (\fcbo), a method for solving the \fcgo problem that models the expectation of the target variable under each policy scope with a Gaussian process model whose inputs are defined in a reproducing kernel Hilbert space. 
\item We validate \fcbo in a synthetic and in a real-world setting with respect to target effects, conditional target effects, and costs of interventions.
\end{itemize}

\section{Background and Setting}\label{sec:background}
We consider a system of observable random variables $\V$ with \emph{target variable} $Y\in \V$ and  \emph{intervenable variables} $\I\subseteq \V\backslash Y$, and the problem of finding a subset of $\I$ and interventions on it that optimize the expectation of $Y$. Our goal is to introduce a method that allows two types of interventions on a variable $X\in\I$: (i) the \emph{hard intervention} $\doi(X=x)$ consisting in setting $X$ to value $x$; and (ii) the \emph{functional intervention}\footnote{Functional interventions are also called \emph{conditional interventions} in \citet{correa2020calculus, correa2020general}.} $X=\pi_{X|\C_X}(\C_X)$ that makes $X$ a deterministic function of a set of variables $\C_X \subseteq \V \backslash \{X, Y\}$, called the \emph{context} of $X$, where  $\pi_{X|\C_X}\colon \range_{\C_X} \mapsto \range_X$ with \eg $\range_{\C_X}$ indicating the range of $\C_X$. Both hard and functional interventions make $X$ a deterministic function of a context $\C_X$ (the hard intervention $\doi(X=x)$ can be viewed as a functional intervention with empty context $\C_X=\emptyset$, setting $X$ to value $x=\pi_{X|\emptyset}(\emptyset)$ where $\pi_{X|\emptyset}\colon \emptyset \mapsto x$ is the empty function), and are therefore referred to as \emph{deterministic interventions} \cite{lee2020characterizing}.

We specify the system's behavior using a \emph{structural causal model} (\scm) ${\cal M}$ defined by the tuple $\langle \V, \U, \calF, p(\U) \rangle$, where $\U$ is a set of exogenous, mutually-independent, unobserved random variables with distribution $p(\U)$, and $\calF=\{f_V\}_{V\in \V}$ is a set of deterministic functions such that $V=f_{V}(\opa(V), \U_V)$ with  $\opa(V) \subseteq \V \backslash V$ and $\U_V \subseteq \U$, $\forall V\in \V$. A deterministic intervention on $X$ therefore replaces $f_X$ with $\pi_{X|\C_X}$.

${\cal M}$ has associated a \emph{directed graph}, which we assume to be \emph{acyclic}\footnote{A directed graph is acyclic if it has no \emph{directed paths} starting and ending at the same node. A directed path is a sequence of linked nodes whose edges are directed and point from preceding towards following nodes in the sequence.}, with nodes $\V \cup \U$ and with an edge from $A$ to $B$ if $A\in \opa(B)$ or $A\in \U_B$. 
A node $A$ with an edge into $B$ is called a \emph{parent} or \emph{direct cause} of $B$ (in this case $B$ is called a \emph{child} of $A$). A node $A$ with a directed path ending at $B$ is called an \emph{ancestor} of $B$ (in this case $B$ is called a \emph{descendant} of $A$). 
We consider the projection of this graph into the graph that contains only nodes $\V$ and that has a directed edge from $V$ to $W$ if $V$ is a parent of $W$ and a bi-directed edge between $V$ and $W$ if $\U_{V} \cap \U_{W} \neq \emptyset$ ($\U_{V} \cap \U_{W}$ is an unobserved \emph{confounder} between $V$ and $W$), and refer to it as \emph{causal graph} associated with ${\cal M}$. Given a causal graph $\graph$, we say that ${\cal M}$ is \emph{compatible} with $\graph$ if all edges that are in the causal graph associated with ${\cal M}$ are also in $\graph$.
We indicate the set of parents, ancestors, and descendants of $V$ in $\graph$ with $\pa_{\graph}(V)$, $\an_{\graph}(V)$ and $\de_{\graph}(V)$, respectively. We indicate the nodes connected to $V$ by a bi-directed edge with $\spo_{\graph}(V)$. We refer to the joint distribution of $\V$ determined by $p(\U)$, which we denote by $p(\V)$, as \emph{observational distribution}.

The space of deterministic interventions for a casual graph $\graph$ can be formalized using the concepts of \emph{mixed policy scope} (\mps) and \emph{deterministic mixed policy} (\dmp) introduced in \citet{lee2020characterizing}. 

\begin{definition}[Mixed Policy Scope (\mps)] \label{def:mps}
A mixed policy scope $\Sset$ for a causal graph $\graph$ 
is a collection of pairs $\langerangle{X}{\C_X}$ such that (i) $X \in \I$, $\C_X \subseteq \V \backslash \{X, Y\}$;
and (ii) the graph $\graph_{\Sset}$ obtained by removing from $\graph$ the incoming edges into $X$ and by adding to $\graph$ directed edges from $\C_X$ to $X$, for every $\langerangle{X}{\C_X} \in \Sset$, is acyclic.
\end{definition}
An \mps specifies the variables in $\I$ on which interventions are performed and their contexts. For example, \mps $\Sset =\{\langerangle{\Aspirin}{\emptyset},\langerangle{\Statin}{\{\Age, \bmi\}}\}$
for $\graph$ in \figref{fig:causalgraphs1}(left) specifies that interventions are performed on $\Aspirin$ and $\Statin$, and with context $\emptyset$ and $\{\Age, \bmi\}$ respectively, as graphically represented in \figref{fig:causalgraphs1}(right). 

\begin{definition}[Deterministic Mixed Policy (\dmp)] \label{def:mps}
A deterministic mixed policy $\piS$ compatible with \mps $\Sset$ is defined as $\piS = \{\pi_{X|\C_X}\}_{\langle X, \C_X \rangle \in \Sset \backslash\Sset_{\hardsubscript}}\bigcup \{\pi_{X|\emptyset }(\emptyset)\}_{\langle X, \C_X \rangle \in \Sset_{\hardsubscript}}$, where $\pi_{X|\C_X}\colon \range_{\C_X} \mapsto \range_X$,  $\pi_{X|\emptyset }(\emptyset)$ denotes the value returned by the empty function, and $\Sset_{\hardsubscript}=\{\langerangle{X}{\C_X}\in \Sset \colon\C_X=\emptyset\}$.
\end{definition}
A \dmp specifies the function $\pi_{X|\C_X}$ or the value $\pi_{X|\emptyset }(\emptyset)$
that replaces $f_X\in \calF$ in ${\cal M}$,
$\forall \langle X, \C_X \rangle \in \Sset$. 
The replacements induce a variant ${\cal M}_{\piS}$ of ${\cal M}$ with joint distribution over $\V$ denoted by $p_{\piS}(\V)$.
We refer to $p_{\piS}(\V)$ as \emph{interventional distribution} induced by $\piS$, and to an observation from $p_{\piS}(\V)$ as an \emph{interventional data sample}.

\section{\fcgo Problem}\label{sec:fcgo}
Let $\mu^Y_{\piS}=\mathbb{E}_{p_{\piS}}[Y]$ denote the expectation of $Y$ w.r.t. the interventional distribution induced by $\piS$, which we refer to as the \emph{target effect}. Our goal is to introduce a method for solving the problem of minimizing $\mu^Y_{\piS}$ over the space $\Sigma$ of \mps{s} for $\graph$ and the space $\PiS$ of \dmp{s} that are compatible with \mps $\Sset$, formally defined below. 
\begin{definition}(\fcgo problem)\label{def:fcgo} 
The \emph{functional causal global optimization} (\fcgo) problem is the problem of identifying a tuple $(\Sset^*,\pi^\st_{\Sset^\st})$ such that 
\begin{align}
\Sset^\st, \pi^\st_{\Sset^\st} & = \argmin_{\Sset \in \Sigma, \piS \in \PiS} \mu^Y_{\piS}\,.
\label{eq:fcgo}
\end{align}
\end{definition}
Importantly, Proposition 1 in \cite{lee2020characterizing} implies that the target effect $\mu^Y_{\pi^\st_{\Sset^\st}}$ given by a solution of the \fcgo problem $(\Sset^*,\pi^\st_{\Sset^\st})$ equals the one that would be obtained by also considering
\emph{stochastic interventions} \citep{correa2020calculus}.

The \fcgo problem extends the \emph{causal global optimization} (\cgo) problem defined in \citet{aglietti2020causal} that only considers hard interventions. In \secref{sec:optimality} we introduce graphical criteria that establish when only considering hard interventions might lead to a bigger target effect and when this is not the case. In addition, in \secref{sec:subgroup} we introduce conditions under which a policy solving the \fcgo problem is also optimal for conditional target effects.

Solving the \fcgo problem requires computing distances between functions defined over different contexts. In \secref{sec:gpsurrogate} we propose to model each target effect via a Gaussian process whose kernel allows computing such distances. We discuss how this approach enables us to keep the computational tractability of standard Bayesian optimization (\bo) methods while allowing to flexibly specify functional interventions.

\subsection{Hard Interventions (Sub-)optimality}\label{sec:optimality}
Let $\Sigma_{\hardsubscript}$ denote the set of \mps{s} in $\Sigma$ that contain only hard interventions, \ie $\Sigma_{\hardsubscript}=\{\Sset\in \Sigma \colon \Sset=\Sset_{\hardsubscript}\}$.
In this section, we introduce graphical criteria that establish when restricting the search space in the \fcgo problem from $\Sigma$ to $\Sigma_{\hardsubscript}$ might lead to a bigger target effect and when this is not the case, thereby informing the investigator about when functional interventions should be considered. The proofs are given in Appendix \ref{sec:app:proofs}.

\begin{proposition}[Sub-optimality of hard interventions]\label{prop:hard_suboptimality}
Let $\graph$ be a causal graph such that (i) $\exists C\in \pa_{\graph}(Y)$ with $C\notin \I$; or (ii) $\exists C\in \text{sp}_{\graph}(Y)$. If $\exists X \in \an_{\graph}(Y) \cap \I$ such that $\{\langle X, C \rangle\}$ is an \mps, then there exists at least one \scm compatible with $\graph$ for which $\min_{\Sset \in \Sigma_{\hardsubscript}, \piS \in \PiS} \mu^Y_{\piS}>\min_{\Sset \in \Sigma, \piS \in \PiS} \mu^Y_{\piS}$.
\end{proposition}

\begin{wrapfigure}[9]{r}{0.12\textwidth}
\vskip-0.5cm
\scalebox{0.8}{
\hskip-0.4cm
\begin{tikzpicture}[dgraph]
    \node[] (a) [label= north:(i)] at (1.7,0.8) {};
    \node[dot] (X) [fill=gray!70,label=south:$X$] at (2, 0) {};
    \node[dot] (C) [fill=darkGreen!70,label=north:$C$] at (3, 1) {};
    \node[dot] (Y) [fill=red!70,label=south:$Y$] at (4, 0) {};
    \draw[line width=0.6pt, \arr](C)--(X);
    \draw[line width=0.6pt, \arr](X)--(Y);
    \draw[line width=0.6pt, \arr](C)--(Y);
\end{tikzpicture}}
\scalebox{0.8}{
\hskip-0.4cm
\begin{tikzpicture}[dgraph]
    \node[] (b) [label= north:(ii)] at (1.7,0.8) {};
    \node[dot] (X) [fill=gray!70,label=south:$X$] at (2, 0) {};
    \node[dot] (C) [fill=gray!70,label=north:$C$] at (3, 1) {};
    \node[dot] (Y) [fill=red!70,label=south:$Y$] at (4, 0) {};
    \draw[{Latex[length=2.mm,width=2.mm]}-{Latex[length=2.mm,width=2.mm]}, dashed, line width=0.6pt](C) to [bend left=+40](Y);
    \draw[line width=0.6pt, \arr](X)--(Y);
    \draw[line width=0.6pt, \arr](C)--(X);
\end{tikzpicture}}
\end{wrapfigure}

\begin{proposition}[Optimality of hard interventions]\label{prop:hard_optimality}
In a casual graph $\graph$, if $\pa_{\graph}(Y)\subseteq \I$ and $\text{sp}_{\graph}(Y)=\emptyset$ there exists a \dmp compatible with \mps $\Sset=\{\langerangle{X}{\emptyset}: X \in \pa_{\graph}(Y)\}$ that solves the \fcgo problem.
\end{proposition}

Proposition \ref{prop:hard_suboptimality} captures two conditions for sub-optimality of hard interventions: the existence of a non-intervenable variable $C$ in $\pa_{\graph}(Y)$ that can serve as context for a functional intervention on a variable $X$, as in the causal graph (i) on the right (for which $\I=\{X\}$); and the existence of a variable $C$ with an unobserved confounder between it and $Y$ that can serve as context for a functional intervention on a variable $X$, as in the casual graph (ii) on the right (for which $\I=\{X, C\}$). In both cases, a hard intervention on $X$ would cut the paths from $X$ to $Y$ passing through $C$ (\ie $X\leftarrow C \rightarrow Y$ and $X\leftarrow C \leftrightarrow Y$ respectively). Instead, a functional intervention on $X$ with context $C$ would keep such paths open and therefore could assign intervention values to $X$ informed by values of $C$, potentially leading to a smaller target effect. Below, we provide two \scm{s} and functional interventions for which this is the case. 

Consider graph (i), with \scm $\mathcal{M}$ with $\U = \{U_C, U_X, U_Y\}$ such that $p(U_C) = p(U_X) = \mathcal{N}(0,1)$ and $p(U_Y) = \mathcal{N}(1, 1)$, and functional assignments $C = U_{C}, X = CU_X, Y = C X U_Y$. $\Sigma_{\hardsubscript}= \{\Sset^1=\{\langerangle{X}{\emptyset}\}\}$ with \dmp $\pi_{\Sset^1}=\{x=\pi_{X|\emptyset}(\emptyset)\}$ induces the modified \scm ${\cal M}_{\pi_{\Sset^1}}$ where $Y = U_C xU_Y$ and $\mu^Y_{\pi_{\Sset^1}} = 0$.
In contrast, \mps $\Sset = \{\langerangle{X}{C}\}$ with \dmp $\pi_{\Sset}=\{\pi_{X|C}(C)=-1/C\}$ induces ${\cal M}_{\piS}$ with $Y = -U_Y$, giving $\mu^Y_{\piS} = -1.0$. Therefore, $\pi_{\Sset}$ achieves a smaller target effect than $\pi_{\Sset^1}$. 

Consider graph (ii), with \scm $\mathcal{M}$ with $\U = \{U_{CY}, U_X, U_Y\}$ such that $p(U_{CY}) = p(U_X) = \mathcal{N}(0, 1)$ and $p(U_Y) = \mathcal{N}(1, 1)$, and functional assignments $C =U_{CY}, X = CU_X, Y = U_{CY} X U_Y$. In this case,   $\Sigma_{\hardsubscript}=\{\Sset^1=\{\langerangle{X}{\emptyset}\},\Sset^2=\{\langerangle{C}{\emptyset}\},\Sset^3=\{\langerangle{X}{\emptyset},\langerangle{C}{\emptyset}\}\}$
with \dmp{s}
$\pi_{\Sset^1}=\{x=\pi_{X|\emptyset}(\emptyset)\}$, 
$\pi_{\Sset^2}=\{c=\pi_{C|\emptyset}(\emptyset)\}$, and  $\pi_{\Sset^3}=\{x=\pi_{X|\emptyset}(\emptyset), c=\pi_{C|\emptyset}(\emptyset)\}$. In ${\cal M}_{\pi_{\Sset^1}}$, $Y =  x U_{CY} U_{Y}$ thus $\mu^Y_{\pi_{\Sset^1}} = 0$. In ${\cal M}_{\pi_{\Sset^2}}$, $Y = cU_X U_{CY}U_{Y}$ thus $\mu^Y_{\pi_{\Sset^2}} = 0$. In ${\cal M}_{\pi_{\Sset^3}}$, $Y = xU_{CY}U_{Y}$ thus  $\mu^Y_{\pi_{\Sset^3}} = 0$.
In contrast, \mps $\Sset = \{\langerangle{X}{C}\}$ with \dmp $\pi_{\Sset}=\{\pi_{X|C}(C) = -1/C\}$ induces ${\cal M}_{\piS}$ with $Y = -U_Y$ giving $\mu^Y_{\piS} = -1$. Therefore $\pi_{\Sset}$ achieves a smaller target effect than any other \dmp containing only hard interventions. 

\subsection{Conditional Target Effects}\label{sec:subgroup}
In addition to potentially leading to a smaller target effect, considering functional interventions allows to deal with settings in which the investigator might wish to minimize the target effect \emph{conditioned} on a set of variables. For instance, in the health example of \figref{fig:causalgraphs1}(left), the investigator might want to find  interventions minimizing the expectation of \psa in a given population as well as in a specific sub-group made of individuals aged over 65, \ie  $\mu^{\psa}_{\piS, \Age > 65}:=\mathbb{E}_{p_{\piS}}[\psa\cond\Age > 65]$ -- since a high percentage of prostate cancer cases are diagnosed within this sub-group \citep{rawla2019epidemiology} -- while still not negatively affecting individuals of other ages. Such settings can be formalized as wishing to minimize the conditional target effect $\mu^Y_{\piS, \C = \c}=\mathbb{E}_{p_{\piS}}[Y\cond \C=\c]$ for $\C \subset \V \backslash Y$ and $\c\in \range_{\C}$. 

Let $\X_{\Sset}$ denote the intervention variables included in \mps $\Sset$, \ie $\X_{\Sset}=\{X \colon \langerangle{X}{\C_X} \in \Sset\}$, and $\C_X^\Sset$ the context variables in \mps $\Sset$ for an intervention on $X$.
Unlike when considering only hard interventions, the following proposition shows that, under some conditions, a solution of the \fcgo problem also minimizes $\mu^Y_{\piS, \C = \c}$ in a restricted \mps{s} space (the proof is given in Appendix \ref{sec:app:proofs}).

\begin{proposition}
[Optimizing conditional target effects]\label{prop:soft_opt}
If  $\Sset^\st, \pi^{\st}_{\Sset^\st}=\argmin_{\Sset \in \Sigma, \piS\in\PiS}\mu^Y_{\piS}$, 
then $\Sset^\st, \pi^{\st}_{\Sset^\st}=\argmin_{\Sset \in \Sigma^{\C}, \piS\in\PiS}\mu^Y_{\piS,\C=\c}$ $\forall \C \subset \V \backslash Y$ 
such that $\C \cap \text{de}_{\graph}(\I)=\emptyset$ and $\forall\c \in \range_{\C}$ with $\Sigma^{\C} = \{\Sset \in \Sigma: \X_{\Sset} = \X_{\Sset^\st} \text{ and } \{\langerangle{X}{\C_X^{\Sset^\st} \cup \C_X^{\Sset} \cup \C}: X \in \X_{\Sset^\st}\} \text{ is an } \mps\}$.
\end{proposition}

\section{Methodology}\label{sec:methodology}
We propose to solve the \fcgo problem using the \emph{functional causal Bayesian optimization} (\fcbo) method summarized in Algorithm \ref{alg:fcbo}, which assumes known casual graph $\graph$ and continuous variables $\V$. 
\fcbo first reduces the search space from $\Sigma$ to a subset $\mismps$ using the \nrmpsreduce~procedure described in \secref{sec:redundancy}; and then solves the minimization problem in \eqref{eq:fcgo} using a Gaussian process (\gptext) $g_{\Sset}(\piS)$ to model the unknown target effect $\mu^Y_{\piS}$, $\forall \Sset \in \mismps$, as described in \secref{sec:gpsurrogate}, with the following sequential strategy. At each trial $t=1,\ldots,T$: (1) \mps $\Sset_t$ and \dmp $\pi^t_{\Sset_t}$ are selected via the expected improvement acquisition functional (\fei) described in \secref{sec:ei}; (2-3) a set of $S$ interventional data samples is obtained and used to compute a sample mean estimate, $\hat{\mu}^Y_{\pi^t_{\Sset_t}}$, of $\mu^Y_{\pi^t_{\Sset_t}}$; (4) $(\pi^t_{\Sset_t}, \hat{\mu}^Y_{\pi^t_{\Sset_t}})$ is added to the interventional dataset $\datai_{\Sset_t}$ of the \mps $\Sset_t$; (5) the posterior distribution of the \gptext $g_{\Sset_t}$,
denoted by $\tau(g_{\Sset_t}\,|\,\mathcal{D}^I_{\Sset_t})$, is updated. Once the maximum number of trials is reached, a tuple  $(\Sset^\st, \pi^\st_{\Sset^\st})$ giving the smallest estimated target effect in $\datai=\{\datai_{\Sset}\}_{\Sset \in \Sigma}$ is returned. 

Notice that Algorithm \ref{alg:fcbo} only requires realizations from $p_{\pi^t_{\Sset_t}}(Y)$ (and could also operate if given directly $\hat{\mu}^Y_{\pi^t_{\Sset_t}}$ instead). This is a considerable practical advantage compared to context-specific reward approaches such as the one in \citet{arsenyan2023contextual} that, similarly to non-causal contextual \bo methods \citep{krause2011contextual}, require values of the contexts and of the target variable resulting from the intervention at that specific context values.  
Similarly to recent approaches in contextual \bo \citep{feng2020high}, \fcbo can directly operate on aggregate rewards.

\begin{algorithm}[t]
\textbf{Inputs:} $\graph$, $\I$, $Y$, $\datai=\{\datai_{\Sset}\}_{\Sset \in \Sigma}$, $T$, $S$\\
$\mismps \leftarrow \nrmpsreduce(\graph, \I, Y)$ \\
Initialise \gpstext~ 
$g_{\Sset}(\piS)$
$\forall\Sset \in \mismps$ with $\datai_{\Sset}$\\
\For{$t=1,\ldots, T$}{
    1. Select \mps $\Sset_t$ and \dmp $\pi^t_{\Sset_t}$ 
    via the \fei
    \\
    2. Obtain samples $\{y^{(s)}\}_{s=1}^{S}$ 
    from $p_{\pi^t_{\Sset_t}}(Y)$\\
    3. Compute sample mean estimate
    $\hat{\mu}^Y_{\pi^t_{\Sset_t}}$ using  $\{y^{(s)}\}_{s=1}^{S}$  
    4. $\datai_{\Sset_t}\leftarrow \datai_{\Sset_t}\cup (\pi^t_{\Sset_t}, \hat{\mu}^Y_{\pi^t_{\Sset_t}})$ \\
    5. Update $\tau(g_{\Sset_t} \cond \datai_{\Sset_t})$
    }
\textbf{Output:}
  $(\Sset^\st, \pi^\st_{\Sset^\st})$ with min $\hat{\mu}^Y_{\pi^\st_{\Sset^\st}}$ over $\datai$
\caption{\fcbo}
\label{alg:fcbo}
\end{algorithm}

\subsection{Search Space Reduction}
\label{sec:redundancy}
The cardinality of $\Sigma$ grows exponentially with the cardinality of $\I$ and the number of possible context sets $\C_X$ for each $X$. Therefore, solving the \fcgo problem by exploring the entire set could be prohibitively expensive.
Even if $\Sigma$ has small cardinality, reducing the search space would simplify the problem by reducing the number of target effects to be modelled. We propose to use the results in  \citet{lee2020characterizing} to reduce the search to the subset of \emph{non-redundant} \mps{s} included in $\Sigma$, denoted by $\mismps$, which is guaranteed to contain a solution to the \fcgo problem. For completeness and clarity, in this section we describe these results in the setting of \dmp{s}.

Let $\Sset^\prime \subseteq \Sset$ indicate that $\C^\prime_{X} \subseteq \C_X$, $\forall \langerangle{X}{\C^\prime_X} \in \Sset^\prime$ with $\X_{\Sset^\prime} \subseteq \X_{\Sset}$. Furthermore, let $\pi_{\Sset^\prime} \subseteq \piS$ indicate that $\pi_{X|\C^\prime_X}(\c_{X}^\prime)=\int \pi_{X|\C_X}(\c^\prime_X \cup \c''_X) p_{\piS}(\c''_X\cond \c^\prime_X) d\c''_X$, $\forall X \in \X_{\Sset^\prime}, \c''_X\in \range_{\C_X\backslash \C^\prime_X}$. Finally, let $\indep_{\graph}$ denote \emph{d-separation} in $\graph$, and $\graph_{\Sset}\backslash X$ the modification of $\graph$ obtained by removing node $X$ and its incoming and outgoing edges.

\begin{definition}[Non-redundant \mps]\label{def:redundant_MPS}
An \mps $\Sset$ is said to be non-redundant if there exists an \scm compatible with $\graph$ and  $\piS \in \PiS$ such that $\mu^Y_{\pi^\prime_{\Sset}} \neq \mu^Y_{\piS}$ $\forall \Sset^\prime \subset \Sset$ and $\pi^\prime_{\Sset} \subset \piS$.
\end{definition}
The following proposition gives a graphical criterion for identifying $\mismps$. 
\begin{proposition}[Characterization of non-redundant \mps]\label{prop:redundancy}
An \mps $\Sset$ is non-redundant if and only if (1) $\X_{\Sset} \subseteq \an_{\graph_\Sset}(Y)$ and (2) $Y \cancel{\indep}_{\graph_{\Sset}\backslash X} C \cond \C_X\backslash C$ for every $X \in \mathbf X_{\Sset}$ and $C \in \C_X$.
\end{proposition}

\subsection{Gaussian Process Surrogate Models}\label{sec:gpsurrogate}
We model the unknown target effect $\mu^Y_{\piS}$ for each $\Sset$ using a \gptext $g_{\Sset}(\piS)$. Differently from existing works on Bayesian functional optimization that focus on univariate functional inputs, $\piS$ can include scalar values as well as functions potentially defined on different input spaces.

\begin{wrapfigure}[9]{r}{0.14\textwidth}
\vskip-0.4cm
\scalebox{0.8}{
\hskip-0.4cm
\begin{tikzpicture}[dgraph]
\node[dot] (c1) [fill=darkGreen!70,label=north:$C_1$] at (-0.8, 2) {};
\node[dot] (c2) [fill=darkGreen!70,label=north:$C_2$] at (0.8, 2) {};
\node[dotdot node] (x) [fill=brightBlue!70,label=north:$X$] at (0, 1.4) {};
\node[dotdot node] (z)[fill=brightBlue!70,label=north:$Z$] at (1.6,1.4) {};
\node[dot] (y) [fill=red!70,label=north:$Y$] at (0.8,0.8) {};
\draw[line width=0.6pt, brightBlue, \arr](c1)--(x);
\draw[line width=0.6pt, brightBlue, \arr](c2)--(x);
\draw[line width=0.6pt, brightBlue, \arr](c2)--(z);
\draw[line width=0.6pt, \arr](x)--(y);
\draw[line width=0.6pt, \arr](z)--(y);
\end{tikzpicture}}
\vskip0.1cm
\scalebox{0.8}{
\hskip-0.4cm
\begin{tikzpicture}[dgraph]
\node[dot] (c1) [fill=darkGreen!70,label=north:$C_1$] at (-0.8, 2) {};
\node[dot] (c2) [fill=darkGreen!70,label=north:$C_2$] at (0.8, 2) {};
\node[dotdot node] (x) [fill=brightBlue!70,label=north:$X$] at (0, 1.4) {};
\node[dotdot node] (z)[fill=brightBlue!70,label=north:$Z$] at (1.6,1.4) {};
\node[dot] (y) [fill=red!70,label=north:$Y$] at (0.8,0.8) {};
\draw[line width=0.6pt, brightBlue, \arr](c1)--(x);
\draw[line width=0.6pt, brightBlue, \arr](c2)--(z);
\draw[line width=0.6pt, \arr](x)--(y);
\draw[line width=0.6pt, \arr](z)--(y);
\end{tikzpicture}}
\end{wrapfigure}
For instance, for \mps $\Sset = \{\langerangle{X}{\{C_1, C_2\}}, \langerangle{Z}{
\{C_2\}}\}$ with $\graph_\Sset$ given on the top right, $\pi_{X|\{C_1, C_2\}}$ is defined over $\range_{C_1} \times \range_{C_2}$, while $\pi_{Z|C_2}$ over $\range_{C_2}$.
Alternatively, for \mps $\Sset = \{\langerangle{X}{\{C_1\}}, \langerangle{Z}{\{C_2\}}\}$ with $\graph_\Sset$ given on the bottom right, $\pi_{X|C_1}$ is defined over $\range_{C_1}$, while $\pi_{Z|C_2}$ over $\range_{C_2}$. 
We address this complexity by introducing a kernel function for  $g_{\Sset}(\piS)$ that allows to compute distances among the mixed inputs while handling the different input dimensionality.

More specifically, $g_\Sset(\pi) \sim \mathcal{GP}(m_\Sset(\pi), K^{\theta}_\Sset(\pi, \pi^\prime))$, where $\pi, \pi^\prime \in\PiS$ (we omit the subscript $\Sset$ to simplify the notation\footnote{In this section, a \dmp $\pi_{\Sset}$ indicates a vector, rather than a set, of interventions.}), and $m_\Sset$ and $K^{\theta}_\Sset$ denote the prior mean and covariance functional with hyperparameters $\theta$. Notice that $\PiS:=\mathbb{R}^{|\Sset_\hardsubscript|} \times \mathcal{B}(\C_{\Sset})$ where\footnote{$|\X|$ indicates the cardinality of the set $\X$.} $\mathbb{R}^{|\Sset_\hardsubscript|}$ is the space of scalar values for $\X_{\Sset_\hardsubscript}$ while $\mathcal{B}(\C_{\Sset})$ is the space of bounded vector-valued functions on $\C_{\Sset}=\bigcup_{X\in \X_{\Sset}}\C_X$.  Given an interventional dataset $\mathcal{D}^I_{\Sset}$ for $\Sset$, for which we assume a Gaussian likelihood, the posterior distribution  $\tau(g_{\Sset}\,|\,\mathcal{D}^I_{\Sset})$ can be computed by standard \gptext updates \citep{williams2006gaussian}.
We initialize $m_\Sset$ to a zero mean functional and extend the \rbf kernel to consider mixed inputs as detailed below.

\textbf{Kernels for Functional \gptext.}
We define $K^{\theta}_\Sset$ as the \rbf kernel $K^{\theta}_S(\pi, \pi')=\sigma^2_f \exp(-||\pi - \pi'||^2/ 2\ell^2)$, where $\theta = (\sigma^2_f, \ell)$ and where
$||\pi - \pi^\prime||$ represents a distance between mixed inputs to the \gptext\footnote{While we discuss the \rbf kernel, this procedure can be used to compute any stationary kernel involving the distance between functional inputs similarly to \cite{vien2018bayesian}.}. Let  $\pi_{\hardsubscript}$ and $\pi_{\softsubscript}$ denote the vectors whose elements are the scalar values and the functions included in $\pi$, respectively. 
We define $||\pi - \pi'||^2$ as $||\pi - \pi'||^2 = ||\pi_{\hardsubscript} - \pi^\prime_{\hardsubscript}||^2 + ||\pi_{\softsubscript} - \pi^\prime_{\softsubscript}||^2_{\mathcal{H}_{\kappa_{\Sset}}}$, with $||\pi_{\hardsubscript} - \pi^\prime_{\hardsubscript}||^2$ indicating the square of the Euclidean distance in $\mathbb{R}^{|\Sset_\hardsubscript|}$, and $||\pi_{\softsubscript} - \pi^\prime_{\softsubscript}||^2_{\mathcal{H}_{\kappa_{\Sset}}}$ the distance between functions in the vector-valued reproducing kernel Hilbert space (\rkhs, \citet{aronszajn1950theory}) $\mathcal{B}(\C_\Sset)=\mathcal{H}_{\kappa_{\Sset}}$ described below. 

Specifically, $\mathcal{H}_{\kappa_{\Sset}}$ is an \rkhs with vector-valued reproducing kernel $\kappa_{\Sset}^{\kappapar}: \range_{\C_{\Sset}} \times \range_{\C_{\Sset}} \to \mathbb{R}^{|\Sset_{\softsubscript}| \times |\Sset_{\softsubscript}|}$ 
where $\kappapar$ denotes the hyper-parameters and $\Sset_{\softsubscript} = \{\langerangle{X}{\C_X} \in \Sset \colon \C_X \neq \emptyset \}$. We refer to $\kappa_{\Sset}^{\kappapar}$ as the \emph{functional intervention kernel} to distinguish it from $K^{\theta}_S$.
We thus have $||\pi_{\softsubscript} - \pi^\prime_{\softsubscript}||^2_{\mathcal{H}_{\kappa_{\Sset}}} = \langerangle{\pi_{\softsubscript}-\pi^\prime_{\softsubscript}}{\pi_{\softsubscript}-\pi^\prime_{\softsubscript}}_{\mathcal{H}_{\kappa_{\Sset}}}$, where  $\langerangle{\cdot}{\cdot}_{\mathcal{H}}$ denotes the inner product in the space $\mathcal{H}$. Evaluating this quantity requires computing $\kappa_{\Sset}^{\kappapar}$ at different input values for the variables in $\C_{\Sset}$, say $\cvalue_{\Sset}$ and $\cvalue^\prime_{\Sset}$, for $\pi_{\softsubscript}$ and $\pi^\prime_{\softsubscript}$ respectively. 

We write the vector of functions $\pi_{\softsubscript}$ included in the \rkhs $\mathcal{H}_{\kappa_{\Sset}}$ as $\pi_{\softsubscript}(\cdot) = \sum_{i = 1}^{N_\alpha} \kappa_{\Sset}^\kappapar(\cvalue_\Sset^i, \cdot)\boldalpha_i$ with $\boldalpha_i \in \mathbb{R}^{|\Sset_{\softsubscript}|}$ and $\cvalue_\Sset^i \in \range_{\C_{\Sset}}$ and let $\pi^\prime_{\softsubscript}(\cdot) = \sum_{i=1}^{N_\beta} \kappa_{\Sset}^{\kappapar}(\cvalue_\Sset^i, \cdot)\mathbf{\boldbeta}_i$ with $\boldbeta_i \in \mathbb{R}^{|\Sset_\softsubscript|}$. This implies that the inner product $\langerangle{\pi_{\softsubscript}-\pi^\prime_{\softsubscript}}{\pi_{\softsubscript}-\pi^\prime_{\softsubscript}}_{\mathcal{H}_{\kappa_{\Sset}}}$ can be written as
\begin{align*}
    &\sum_{i=1}^{N_\alpha} \sum_{j=1}^{N_\alpha} \boldalpha_i^\top \kappa_{\Sset}^{\kappapar}(\cvalue^i_\Sset, \cvalue^j_\Sset) \boldalpha_j
    +\sum_{i=1}^{N_\beta} \sum_{j=1}^{N_\beta}  \boldbeta_i^\top \kappa_{\Sset}^{\kappapar}(\cvalue^i_\Sset, \cvalue^j_\Sset) \boldbeta_j \\ &- 2 \sum_{i=1}^{N_\alpha} \sum_{j=1}^{N_\beta} \boldalpha_i^\top \kappa_{\Sset}^{\kappapar}(\cvalue^i_\Sset, \cvalue^j_\Sset) \boldbeta_j.
\end{align*}
To construct $\kappa_{\Sset}^\kappapar$, we propose to augment the input space 
by including a task index for each function $\pi_{X|\C_X}$ in $\Sset$, \ie we redefine $\kappa_{\Sset}^\kappapar \colon (\range_{\C_\Sset} \times \mathcal{T}) \times (\range_{\C_\Sset} \times \mathcal{T}) \to \mathbb{R}^{|\Sset_\softsubscript| \times |\Sset_\softsubscript|}$ where $\mathcal{T}$ is the space of integer values from 1 to $|\Sset_\softsubscript|$. For every realization of the context variables and the task index, say $(\cvalue_\Sset, t)^i$, we can then evaluate $\kappa_{\Sset}^\kappapar((\cvalue_\Sset, t)^i, (\cvalue_\Sset, t)^j)$. We assume the covariance between functions defined on different input spaces, \ie for which $t^i \neq t^j$, to be 0\footnote{Alternative kernel constructions where this assumption is relaxed are discussed in Appendix \ref{sec:app:alternate_kernels}.}. Instead, we let the covariance structure across function values associated with different inputs for $t^i = t^j$ be determined by a task-specific kernel, which we denote by $k^{t^i}$. Denote by $\cvalue^i_\Sset[t^i]$ the subset of values included in $\cvalue^i_\Sset$ for the contexts of the $t^i$ task and by $\kappapar[t^i]$ the subset of hyper-parameters for $t^i$ included in $\kappapar$. We have that $\kappa_{\Sset}^\kappapar((\cvalue_\Sset, t)^i, (\cvalue_\Sset, t)^j)$ is equal to $k^{t^i}(\cvalue^i_\Sset[t^i], \cvalue^j_\Sset[t^j])$ with hyper-parameters $\kappapar[t^i]$ if $t^i=t^j$ and to 0 otherwise. The kernel $k^{t^i}$ might differ across tasks both in terms of functional form and hyper-parameter values. This allows to impose different characteristics in terms of \eg smoothness for each function $\pi_{X|\C_X}$ included in $\pi$. 

\subsection{Acquisition Functional} \label{sec:ei}
We sequentially select interventions by numerically\footnote{Alternatively, the functional gradient w.r.t. functions in a
\rkhs could be derived analytically (see \cite{vien2018bayesian}).} maximizing the expected improvement (\acro{ei}) per unit of cost $\cost_{\Sset}(\cdot)$ \emph{across} the \mps{s} in $\mismps$. Given an interventional dataset $\datai_{\Sset}$, for each $\Sset \in \mismps$ the functional \acro{ei} (\fei) is given by:
\begin{align*}
\text{\fei}_{\Sset}(\pi) = \sigma^2_\Sset(\pi \cond \mathcal{D}^I_{\Sset})[\gamma(\pi)\boldsymbol{\Phi}(\gamma(\pi)) + \phi(\gamma(\pi))]/\cost_{\Sset}(\pi),
\end{align*}
where $\sigma^2_\Sset(\pi \cond \mathcal{D}^I_{\Sset}) = K^\theta_\Sset(\pi, \pi \cond \mathcal{D}^I_{\Sset})$, $\boldsymbol{\Phi}(\cdot)$ and $\phi(\cdot)$ are the \acro{cdf} and \acro{pdf} of a standard Gaussian random variable respectively, and $\gamma(\pi) = \frac{m_{\Sset}(\pi \cond \mathcal{D}^I_{\Sset}) - g^\st}{K^\theta_\Sset(\pi, \pi \cond \mathcal{D}^I_{\Sset})}$
with $g^\st$ denoting the optimum observed for $g_{\Sset}$ across \mps{s} in $\mismps$. 
$m_{\Sset}(\pi\cond\mathcal{D}^I_{\Sset})$ and $K^\theta_\Sset(\pi, \pi \cond \mathcal{D}^I_{\Sset})$
denote the posterior parameters of $\tau(g_{\Sset}\cond \datai_{\Sset})$. At every trial $t$ of the optimization, the \mps and the \dmp are chosen by numerically solving $\Sset_t, \pi^t_{\Sset_t} = \argmax_{\Sset \in \mismps, \pi_\Sset \in \Pi_{\Sset}} \text{\fei}_{\Sset}(\pi_\Sset)$.

$\cost_{\Sset}(\pi)$ denotes the cost associated to $\pi$.  
We consider two types of costs: (i) $\cost_{\Sset}(\pi) = |\Sset|$; and  (ii) $\cost_{\Sset}(\pi) = \sum_{X \in \X_{\Sset}} \int_{\range_{\C_X}} \pi_{X|\C_X}(\c_X) d\c_X$, (\ie the sum of the area under $\pi_{X|\C_X}$ over all $X\in \X_{\Sset}$), which can be seen as a measure of the units of intervention given to a population whose context values $\c_X$ are uniformly distributed in $\range_{\C_X}$. Notice that the second cost requires knowledge of $\range_{\C_X}$ at initialization. 
We use the first cost in the \echain experiments of \secref{sec:echain}, and the second cost in the \health experiments of \secref{sec:health}.

\section{Related Work}
There exist two other \cbo-type methods in the literature that can achieve contextual interventions, namely \cocabo \citep{arsenyan2023contextual} and \mcbo \citep{sussex2022model}. \cocabo performs different hard interventions in separate sub-groups defined by some contexts \emph{after} observing context values.
Interventional data samples, formed by context values, intervention values, and target effect, are used to fit a \gptext model over the potentially high-dimensional context-intervened variables space. Therefore, \cocabo can only be used in settings in which the investigator observes the values of the context variables, say $\C=\c$, selects an intervention and observes the resulting target effect across units with $\C=\c$, rather than an aggregate target effect across all possible context values in a population. This is not feasible in many applied problems (\eg in \acro{a}/\acro{b} testing platforms, in which outcomes are often measured as an aggregate across a large population that spans an entire distribution of contexts), and might lead to sup-optimal policies for unseen or rarely observed context values. In addition, this method defines the \gptext surrogate model for each \mps $\Sset$ on $\C_{\Sset}$ thus reducing the flexibility of the learned policy by not encoding the existence of different $\C_X$ for each $X$ in $\X_{\Sset}$.
\mcbo considers systems described by \scms~in which $X\in \I$ is of the form $X = f_X(\pa_{\graph}(X), \A_X) + U_X$, where $\A_X$ is a set of action variables that parametrize $f_X$ whose values can be set by the investigator to induce a change in the parametrization. 
Therefore, a contextual intervention in \mcbo \emph{modifies} a node's original functional assignment rather than \emph{replacing} it as in \fcbo. This might lead to more limited interventions and does not allow change of contexts. In addition, this method can achieve contextual interventions only in settings in which the system's \scm contains action variables. When this is not the case, \mcbo can only implement hard interventions (see the \health experiment of \secref{sec:health}). 
Finally, unlike \fcbo, \mcbo does not reduce the search space and cannot handle unobserved confounders. 

Extensions of \bo \citep{shahriari2015taking} to solve functional global optimization (\fgo) problems have been studied by searching over the space of Bernstein polynomials  \citep{vellanki2019bayesian}, by constructing a sequence of low-dimensional search spaces \citep{shilton2020sequential}, or by representing the functional inputs as elements in an \rkhs (\fbo) \citep{vien2018bayesian}. This work takes an approach similar to \fbo, but considers a varied search space and its causal reduction. More importantly, thanks to a simple kernel construction, it enables functional \bo, which has generally focused on univariate functional inputs, to deal with settings where the inputs are multi-task functions.
\begin{figure*}[t]
    \centering
    \includegraphics[width=0.32\textwidth]{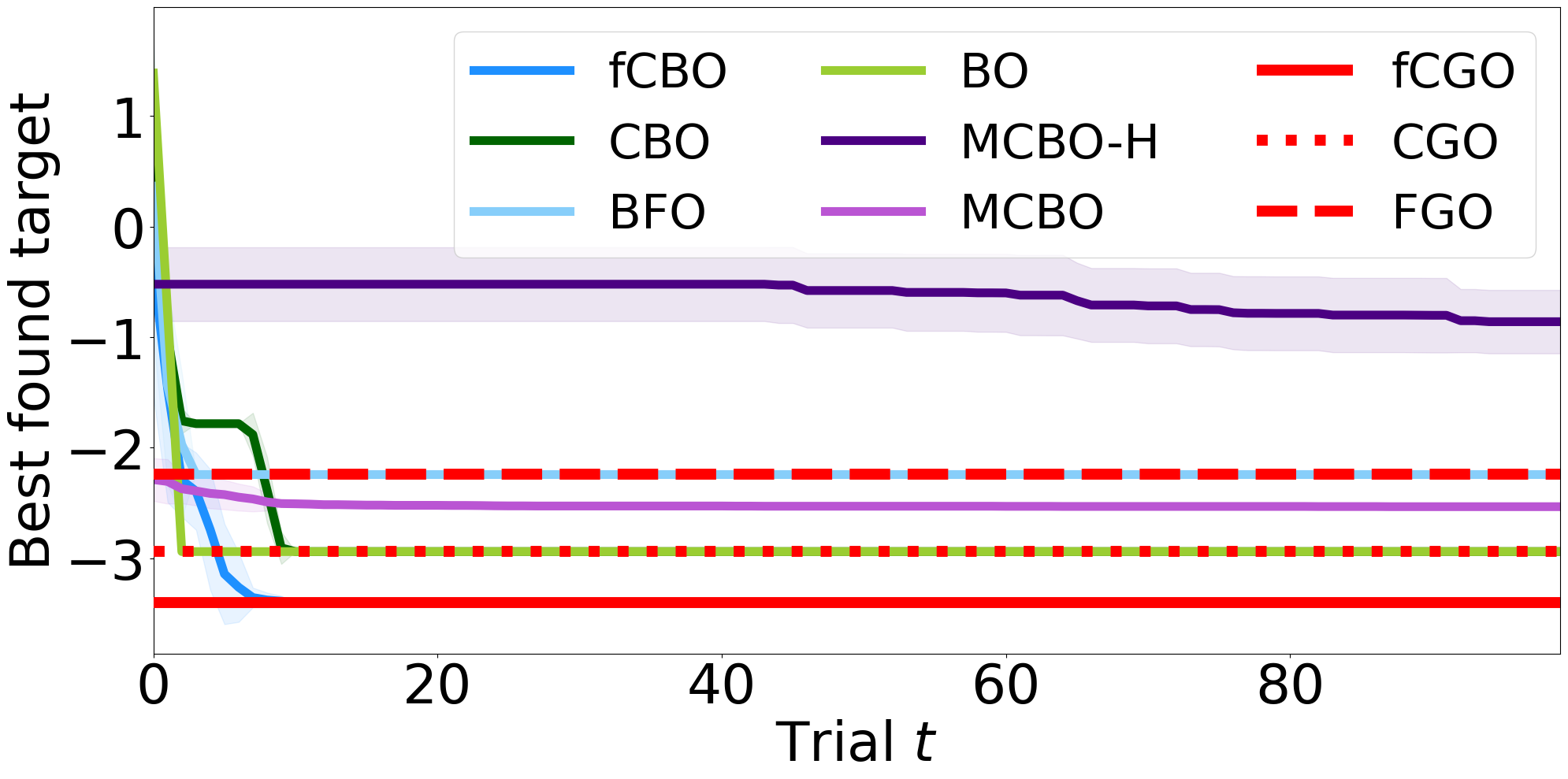}
    \hspace{0.5em}
    \includegraphics[width=0.32\textwidth]{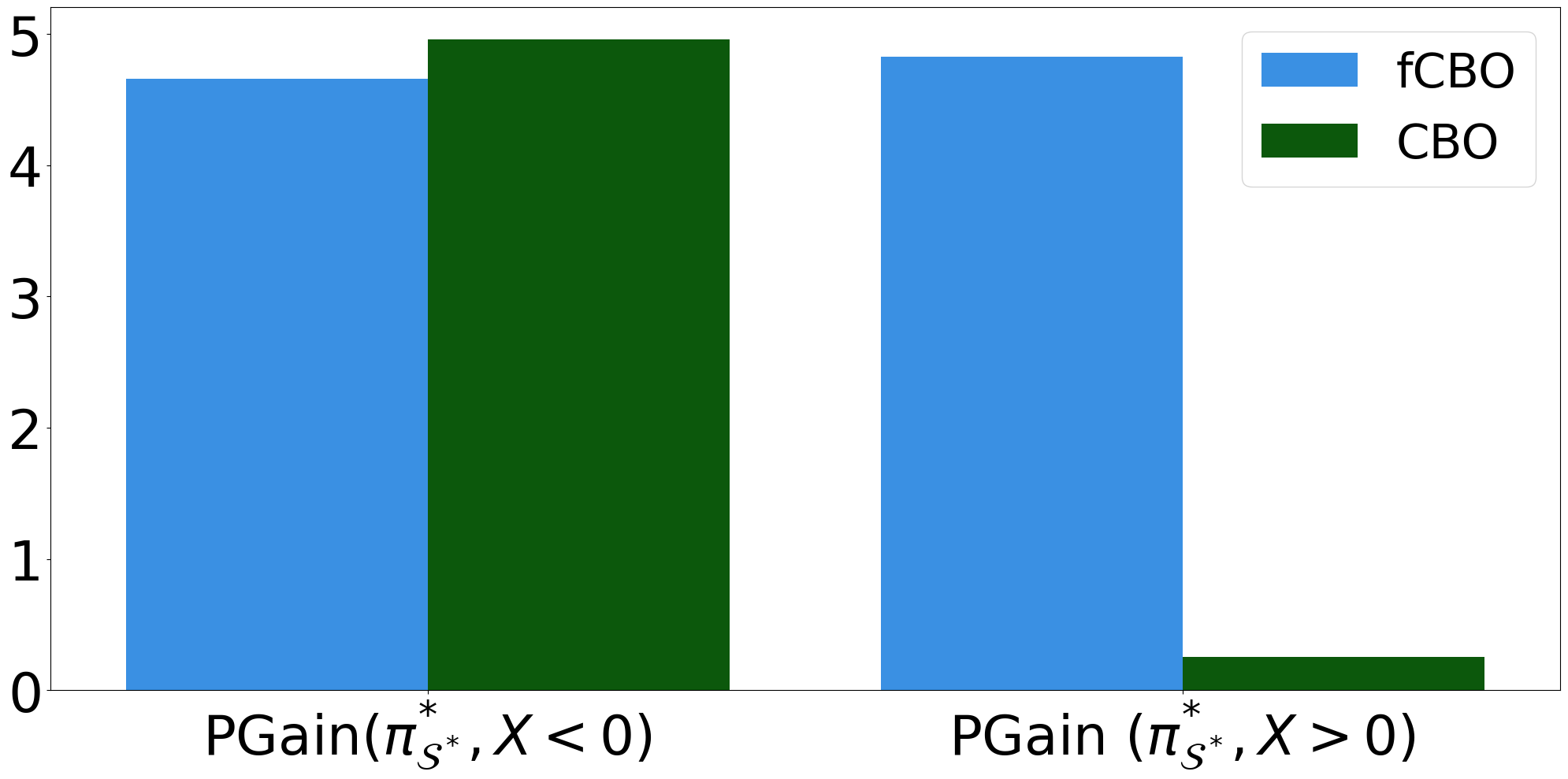}
    \includegraphics[width=0.32\textwidth]{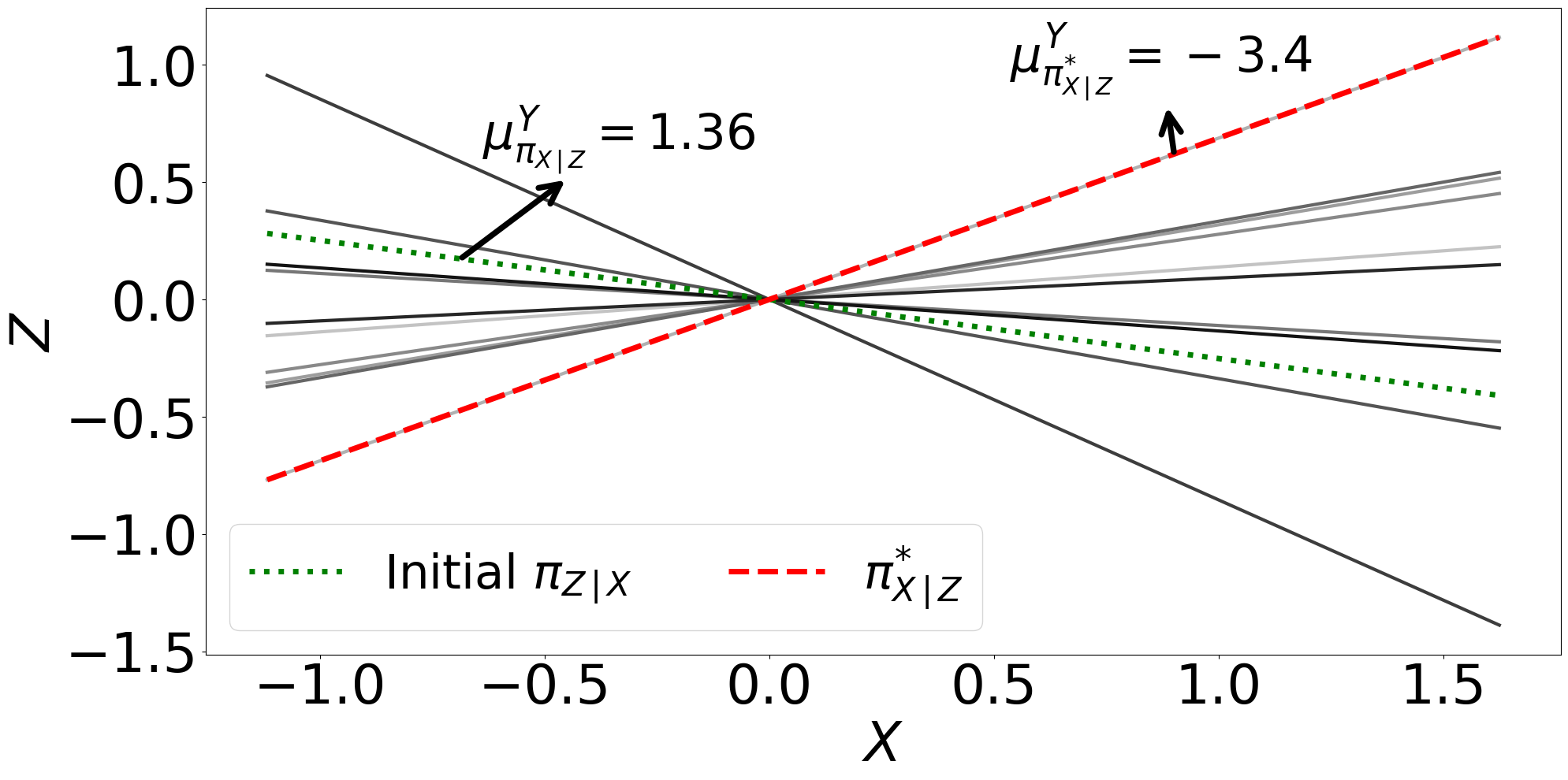}
    \caption{\echain experiments. \emph{Left}: 
    Average convergence of \fcbo to the \fcgo optimum (solid red line); of \cbo,  \bo, and \mcbo-\textsc{h} to the \cgo optimum (dotted red line); of \fbo to the \fgo optimum (dashed red line); and of \mcbo to the $\cgo^*$ optimum (across 20 initializations of $\datai$ for \fcbo, \cbo, \bo, \fbo, and across 20 seeds for \mcbo-\textsc{h} and \mcbo~-- shaded areas give $\pm$ standard deviation).
    \emph{Middle}: Average performance gains $\text{\acronospace{pg}ain}(\pi^\st_{\Sset^\star},X<0)$ and $\text{\acronospace{pg}ain}(\pi^\st_{\Sset^\star},X>0)$ obtained by the optimal \dmp $\pi^\st_{\Sset^\star}$. 
    \emph{Right}: Initial $\pi_{Z|X}$ included in $\datai_Z$, $\pi^*_{Z|X}$ found by \fcbo and associated target effect values.
    }
    \label{fig:aug_chain_full_results}
\end{figure*}
\section{Experiments}\label{sec:experiments}
We compare\footnote{We cannot compare to \cocabo as: (i) in our settings the values of the contexts are not observed before intervening, and only an aggregate target effect across contexts is observed post intervention; (ii) this method does not allow considering \mps{s} that do not share the same contexts.} \fcbo with \cbo, \mcbo, \bo, and \fbo on the synthetic graph in \secref{sec:echain} (\echain), and on 
the healthcare graph in \figref{fig:causalgraphs1}(a) (\health). 
The experiments aim at highlighting three main advantages of using \fcbo to find optimal interventions. The first advantage is the ability to achieve smaller target effects compared to methods that use only hard interventions. We assess this by looking at the convergence to the optimum. The second advantage is the ability to perform well w.r.t. conditional target effects. We demonstrate this in the \echain experiments, by computing the \emph{performance gain} for \dmp $\pi_{\Sset}$ on sub-group $\C=\c$, which is defined as $\text{\acronospace{pg}ain}(\pi_{\Sset},\C=\c) = \hat\mu^Y_{\C=\c} -\hat\mu^Y_{\pi_{\Sset}, \C=\c}$, where $\hat\mu^Y_{\C=\c}$ denotes an estimate of the conditional expectation of $Y$ given $\C=\c$ w.r.t. the observational distribution and $\hat\mu^Y_{\pi_{\Sset}, \C=\c}$ an estimate of the conditional target effect. 
The third advantage is the ability to craft flexible and more targeted \dmp{s} that can incur similar or lower cost, while still ensuring a smaller target effect than policies made of only hard interventions. We exemplify this in the \health experiments where we assume a cost function given by $\cost_{\Sset}(\pi)=\sum_{X \in \X_{\Sset}} \int_{\range_{\C_X}} \pi_{X|\C_X}(\c_X) d\c_X$.\\

\begin{minipage}[c]{1\linewidth}
\resizebox{1\textwidth}{!}{
\begin{tabular}{cccccc}
\multicolumn{6}{c}{\multirow{2}{*}{\textbf{Search Space and Optimization Problem}}} \\  \\
\cellcolor{brightBlue}{\color[HTML]{FFFFFF} \textbf{\fcbo}} & \cellcolor{cboGreen}{\color[HTML]{FFFFFF} \textbf{\cbo}} & \cellcolor{darkPurple}{\color[HTML]{FFFFFF} \textbf{\mcbo-\textsc{h}}} & \cellcolor{mcbosoftpurple}{\color[HTML]{FFFFFF} \textbf{\mcbo}} & \cellcolor{lightGreen}{\color[HTML]{333333} \textbf{\bo}} & \cellcolor[HTML]{DAE8FC}{\color[HTML]{333333} \textbf{\fbo}} \\
\multirow{ 2}{*}{$\Sigma$} & \multirow{ 2}{*}{$\Sigma_{\hardsubscript}$} & \multirow{ 2}{*}{$\Sigma_{\hardsubscript}$} & \multirow{ 2}{*}{$\mathcal{P}_{\boldsymbol{A}}$} & \multirow{ 2}{*}{$\Sset_{\I, \C_X=\emptyset}$} & \multirow{ 2}{*}{$\Sset_{\subseteq\I, \C_X\neq\emptyset}$}\\ \\ \hline 
\multirow{ 2}{*}{\fcgo} & \multirow{ 2}{*}{\cgo} & \multirow{ 2}{*}{\cgo} & \multirow{ 2}{*}{$\cgo^\st$} & \multirow{ 2}{*}{\acro{go}} & \multirow{ 2}{*}{\fgo}\\ \\ \hline
\vspace*{0.01cm}
\end{tabular}
}
\end{minipage}

The different search spaces of \fcbo, \cbo, \mcbo, \bo, and \fbo are summarized in the table above. An intervention in \bo and \fbo is performed on all variables or on a subset of variables in $\I$ simultaneously: \bo considers only hard interventions, thus its search space contains only \mps $\Sset_{\I, \C_X=\emptyset}=\{\langerangle{X}{\C_X}\colon X\in \I, \C_X=\emptyset\}$); while \fbo considers functional interventions with a fixed $\C_X\neq \emptyset$ over trials, \ie its search space contains only one \mps
formed by tuples $\langerangle{X}{\C_X}$ with $\X_\Sset\subseteq \I$, denoted by $\Sset_{\subseteq \I, \C_X\neq\emptyset}$. \cbo and \mcbo with hard interventions, denoted by \mcbo-\textsc{h}, consider the  space of \mps{s} containing only hard interventions $\Sigma_{\hardsubscript}$. Finally, \mcbo performs interventions via actions variables $\boldsymbol{A} = \{\boldsymbol{A}_X\}_{X\in \I}$ thus exploring the power set $\mathcal{P}_{\boldsymbol{A}}$ (with the convention that no intervention on $X$ corresponds to removing $\A_X$ from the \scm). While \fcbo aims at solving the \fcgo problem, \cbo and \mcbo-\textsc{h} target the \cgo problem, and \fbo the \fgo problem. Finally, \bo solves a global optimization problem (\acro{go}), while \mcbo a \cgo problem in the action variable space, denoted by $\cgo^*$. In all experiments, we consider settings where the \fcgo, \cgo, and \fgo problems have unique solutions, and the \acro{go} optimum coincides with the \cgo optimum.
\begin{figure*}[t]
    \centering
    \hspace{-0.2cm}
    \includegraphics[width=0.36\textwidth]{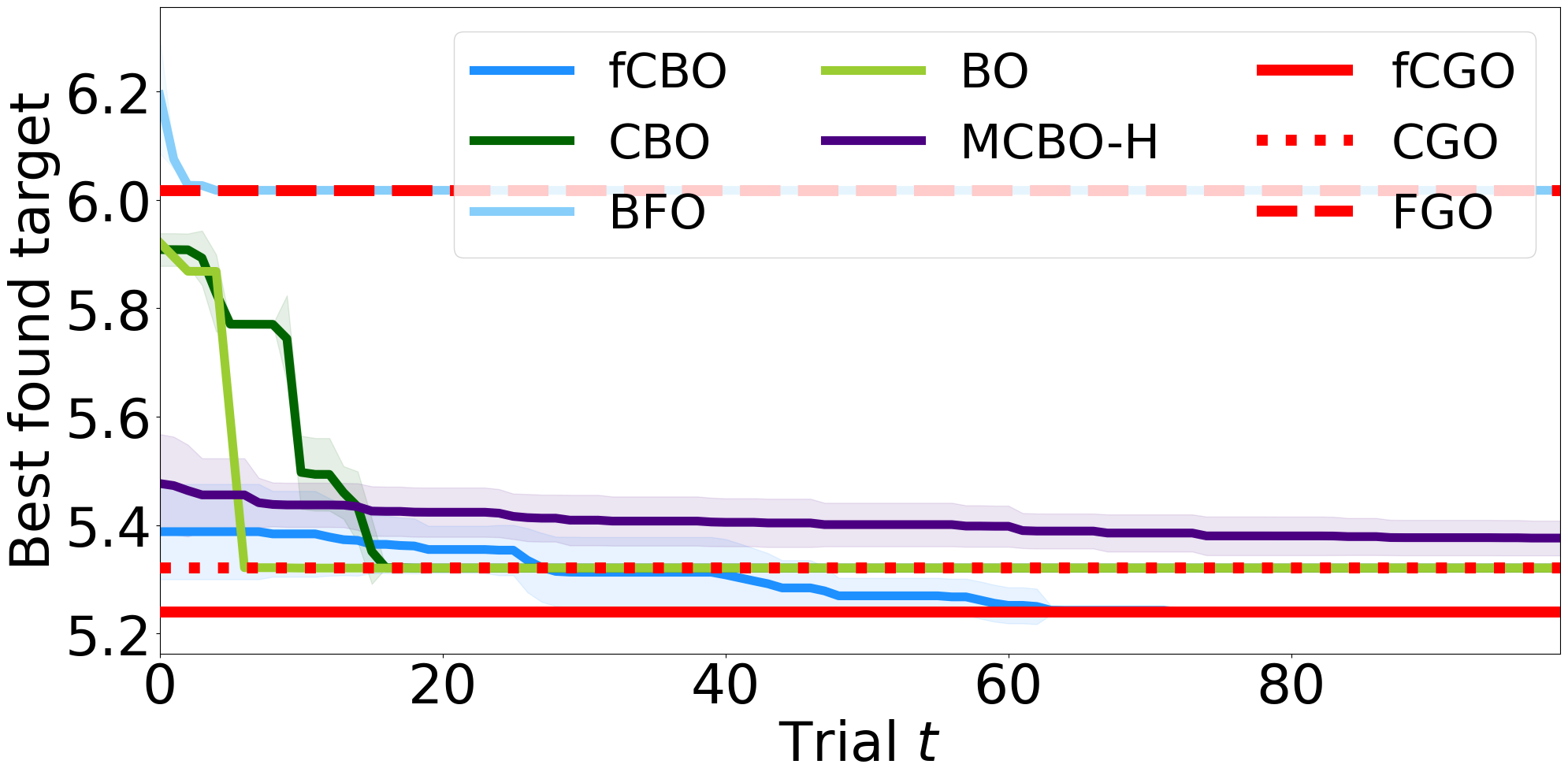}
    \includegraphics[width=0.23\textwidth]{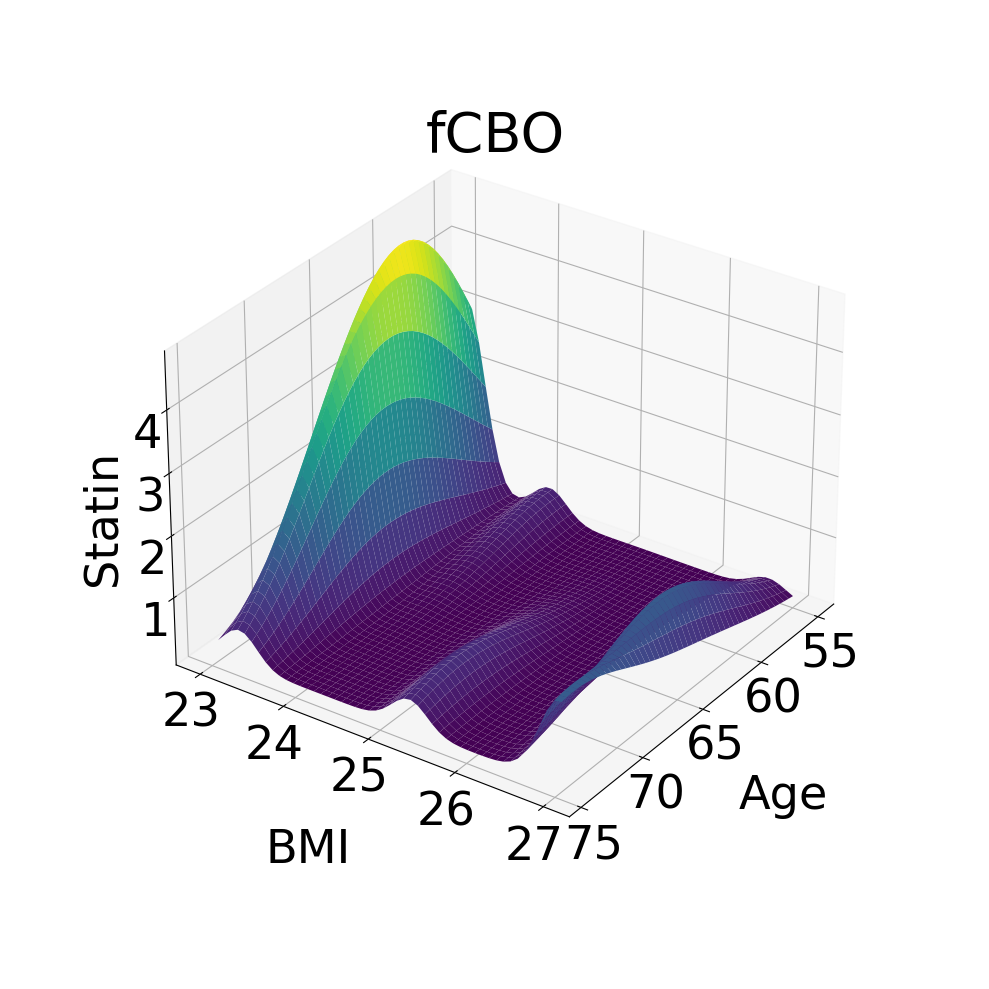}
    \hspace{-0.7cm}
    \includegraphics[width=0.23\textwidth]{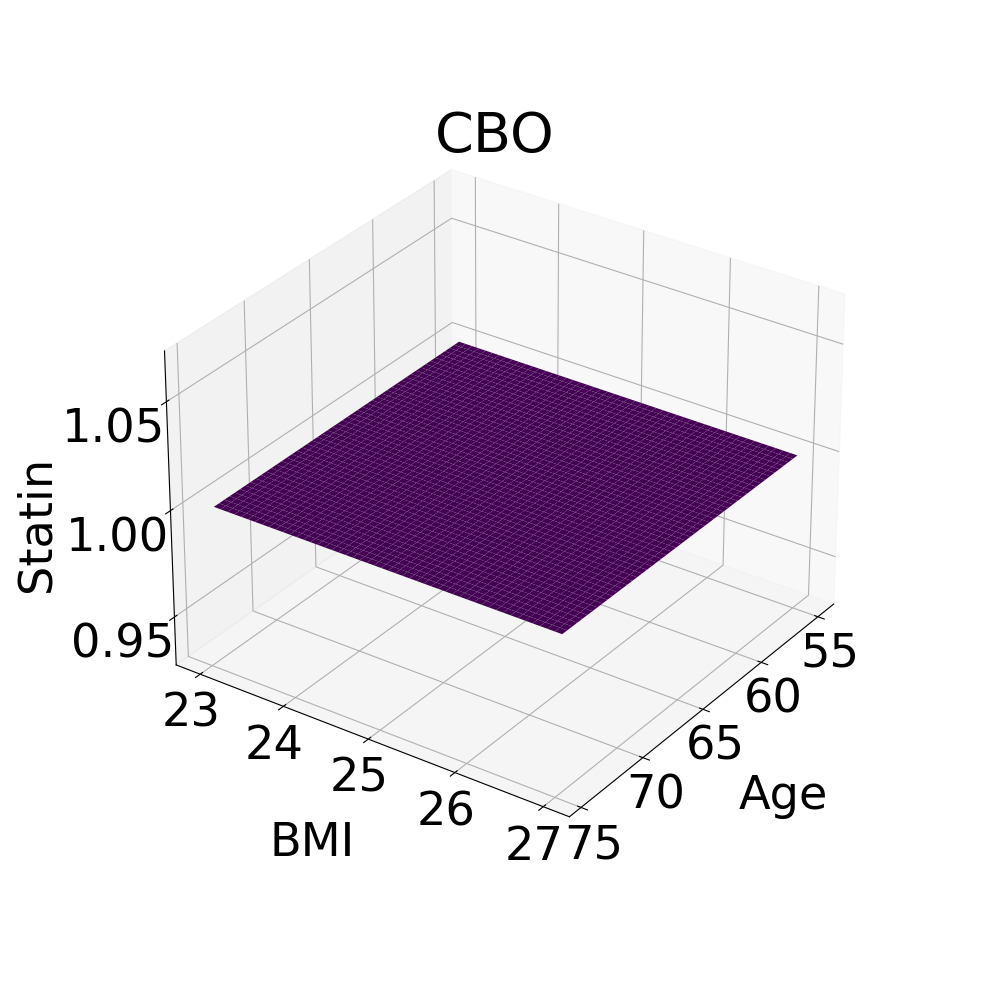}
    \includegraphics[width=3.3cm, height=3.3cm, keepaspectratio]{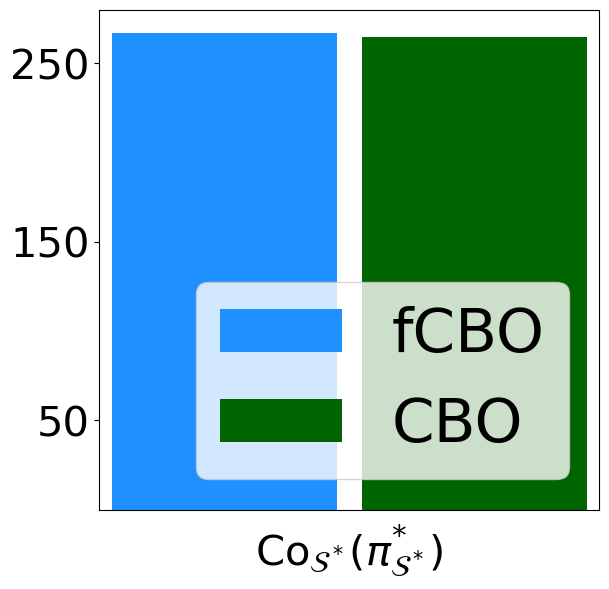}
    \caption{\health experiments with $\texttt{GridSize}=5$. \emph{Left}: Average convergence of \fcbo to the \fcgo optimum (solid red line); of \cbo, \bo, and \mcbo-\textsc{h} to the \cgo optimum (dotted red line); and of \fbo to the \fgo optimum (dashed red line) (across 20 initializations of $\datai$ for \fcbo, \cbo, \bo, \fbo, and across 20 seeds for \mcbo-\textsc{h} -- shaded areas give $\pm$ standard deviation).
    \emph{Middle}: $\pi^*_{\Statin|\Age, \bmi}$ found by \fcbo (left) and $\pi^*_{\Statin|\emptyset}$ found by \cbo (right) across levels of \Age~and \bmi. 
    \emph{Right}: Cost associated to the optimal \mps and associated optimal \dmp found by \fcbo and \cbo.}
    \label{fig:health_full_results}
\end{figure*}

\textbf{\fcbo, \cbo, \bo, and \fbo.}
While \fcbo does not impose restrictions in terms of context variables used for functional interventions beyond acyclicity of $\graph_{\Sset}$, for ease of demonstration and for computational reasons, in the experiments we only consider keeping the original parents as contexts. In other words, we set $\C_X=\pa_{\graph}(X)$ for each functional intervention. We make the same choice for \fbo. To demonstrate performance on different choices for $\PiS$, we consider linear and \rbf functional intervention kernels $\kappa_{\Sset}^{\kappapar}$ in the \echain and \health experiments, respectively. We use the same functional intervention representation for \fbo. 
For each $\Sset \in \mismps$ we numerically optimize the acquisition functions on a grid whose size is set to $\texttt{GridSize}^{|\Sset_{\hardsubscript}| + 1}$ where $\texttt{GridSize}$ is a hyper-parameter. We initialize $\datai$ by randomly generating a single \dmp and associated target effect for each $\Sset \in \Sigma$. We provide average results across the 20 different initializations. 

\paragraph{\mcbo.}
In the \echain experiments, we consider both \mcbo restricted to hard interventions (\mcbo-\textsc{h}) and \mcbo with contextual interventions (by augmenting the \scm with an action variable for each variable in $\I$). In the \health experiments, the \scm is given and does not contain action variables. Therefore, we follow \cite{sussex2022model} and consider only hard interventions on $\Aspirin$, $\Statin$, and $\ci$.  We run the algorithm\footnote{We used the code companion to \cite{sussex2022model} available at \url{https://github.com/ssethz/mcbo}.} by setting the random seed controlling both the initial interventional data and the optimization of the acquisition function to values $1,\ldots,20$. We report results across the 20 different seeds. Cross-validation with values $0.05, 0.5$, and $5$ on the hyper-parameter $\beta$ for the \acro{ucb} acquisition function, as done in \cite{sussex2022model}, does not give major differences in the performance (we report the results for $\beta=5$).

\subsection{\echain Experiments}\label{sec:echain}
\begin{wrapfigure}[8]{r}{0.19\textwidth}
\vskip-0.7cm
\begin{minipage}[c]{.19\linewidth}
\hskip-0.19cm
\scalebox{0.8}{
  \begin{tikzpicture}[dgraph]
    \node[dot] (X) [fill=darkGreen!70,label=north:$X$] at (1.,0) {};
    \node[dot] (Z) [fill=gray!70,label=north:$Z$] at (2.5,0) {};
    \node[dot] (W) [fill=gray!70,label=north:$W$] at (3,1) {};
    \node[dot] (Y) [fill=red!70,label=north:$Y$] at (4,0) {};
    \draw[line width=0.6pt, \arr](X)--(Z);
    \draw[line width=0.6pt, \arr](Z)--(Y);
    \draw[line width=0.6pt, \arr](W)--(Y);  
    \draw[line width=0.6pt, \arr](X)to [bend right=+20](Y);
    \end{tikzpicture}}
\end{minipage}  
\vskip-0.1cm
\begin{minipage}[c]{.15\linewidth}
\begin{small}
\begin{align*}
    & X = U_X, \, W = U_W \\
    & Z = -0.5X + U_Z \\
    & Y = -W -3ZX + U_Y
\end{align*}
\end{small}
\end{minipage}
\end{wrapfigure}
We first experiment on the chain graph with associated \scm given on the right (see Appendix \ref{sec:app:echain} for details). \figref{fig:aug_chain_full_results}(left) shows how considering mixes of hard and functional interventions allows \fcbo to reach the smallest target effect.

\figref{fig:aug_chain_full_results}(middle) shows how \fcbo and \cbo differ in terms of conditional target effects defined for $X<0$ and $X>0$. Due to the existence of the interaction term $-3ZX$, minimizing $Y$ would require setting $Z$ to a negative value when $X<0$ and to a positive value when $X>0$. However, this cannot be achieved via hard interventions that set $Z$ to a fixed value irrespective of $X$ as in \cbo. As a consequence \cbo, which selects \mps $\Sset^{\st} = \{\langle Z, \emptyset \rangle, \langle W, \emptyset \rangle\}$ and \dmp $\pi^\st_{\Sset^\st} = \{-1, 1\}$, achieves a very low performance gain for $X>0$,  $\text{\acronospace{pg}ain}(\pi^\st_{\Sset^\star},X>0)$. Instead,
\fcbo selects \mps $\mathcal{S}^\st = \{\langle Z, X\rangle, \langle W, \emptyset \rangle \}$ and \dmp $\pi^\st_{\Sset^\st}= \{\pi^\st_{Z|X}, 1\}$, where the linear function $\pi^\st_{Z|X}$ (shown as a dashed red line in \figref{fig:aug_chain_full_results}(right)) has a slope that gives an optimal $Z$ value for both sub-groups thus leading to an evenly distributed performance gain.

\subsection{\health Experiments}\label{sec:health}
\vskip-0.2cm
For the \health experiments, we use the \scm by \citet{ferro2015use} (see Appendix \ref{sec:app:health} for details).
\figref{fig:health_full_results} shows the results obtained with $\texttt{GridSize}=5$.
In these experiments, \fcbo achieves the smallest target effect by selecting \mps $\Sset^\st = \{\langle \Aspirin, \emptyset \rangle, \langle \Statin, (\Age, \bmi) \rangle, \langle \ci, \emptyset \rangle\}$ and \dmp $\pi^\st_{\Sset^\st} = \{ 0.1, \pi^\st_{\Statin|\Age,\bmi}, 1\}$. \bo and \cbo select \mps $\Sset^\st = \{\langle \Aspirin, \emptyset \rangle, \langle \Statin, \emptyset \rangle, \langle \ci, \emptyset \rangle\}$, and \dmp $\pi^\st_{\Sset^\st} = \{0.1, 1, 1\}$. \mcbo-\textsc{h} does not reach convergence. 

\figref{fig:health_full_results}(middle) displays $\pi^\st_{\Statin|\Age,\bmi}$ selected by \fcbo (left) and $\pi^\st_{\Statin|\emptyset}(\emptyset)=1$ selected by \cbo as a constant function over \Age~and \bmi (right). These two plots show that, while methods that consider only hard interventions are forced to assign intervention values uniformly across the context space, methods that also allow functional interventions can concentrate on specific sub-groups, in this case characterized by lower values of Age and \bmi. Being able to differentiate among interventions assigned to different sub-groups has important implications in terms of cost $\cost_{\Sset^\st}(\pi^\st_{\Sset^\st})$.
\figref{fig:health_full_results}(right) shows that \fcbo incurs almost the same cost as \cbo. This result demonstrates another key property of functional interventions: taking the context values into account allows the investigator to assign interventions to units in the population characterized by context values that lead to smaller target effects. 
\begin{figure*}[t]
    \centering
    \hspace{-0.2cm}
    \includegraphics[width=0.36\textwidth]{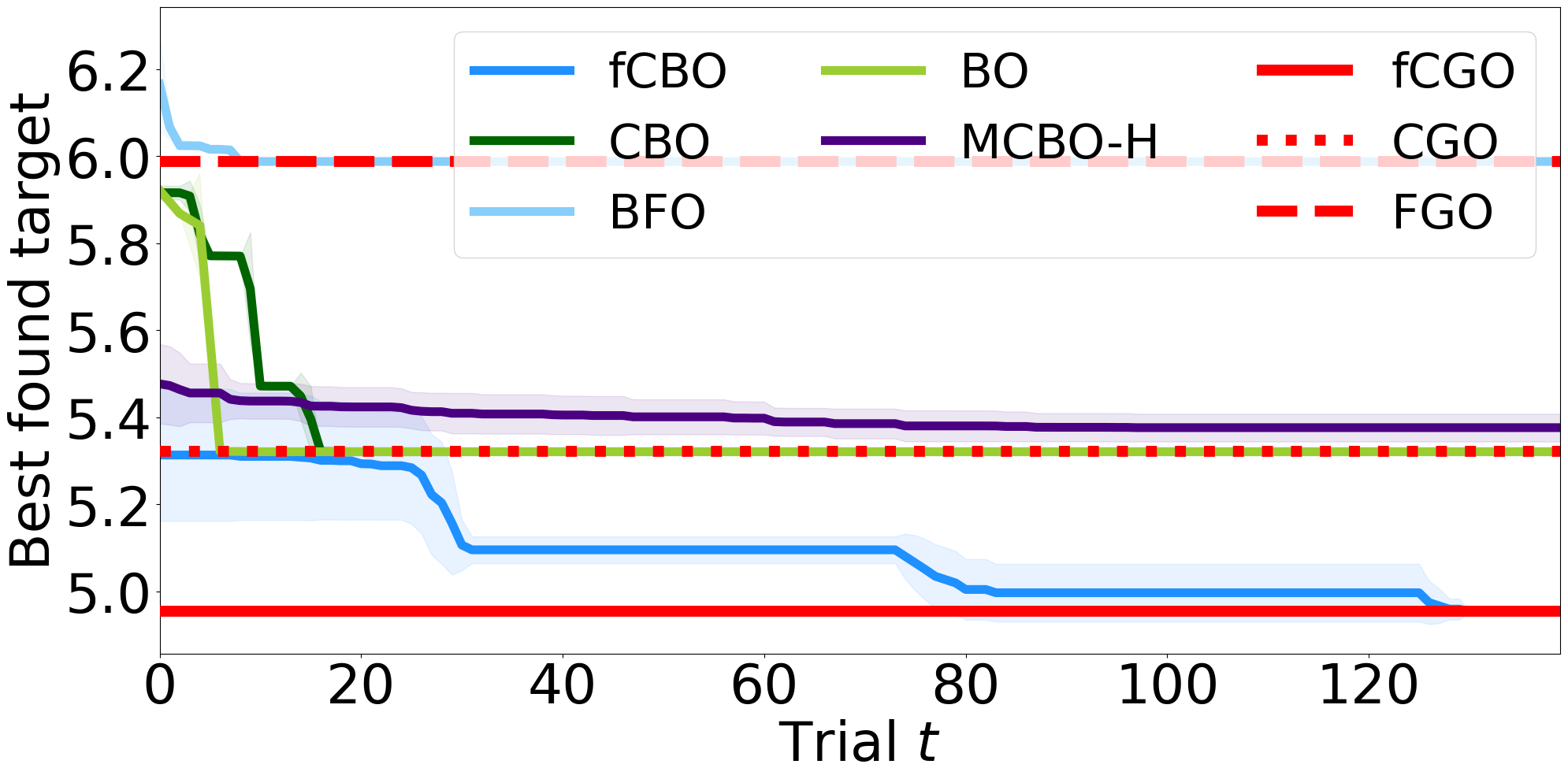}
    \includegraphics[width=0.23\textwidth]{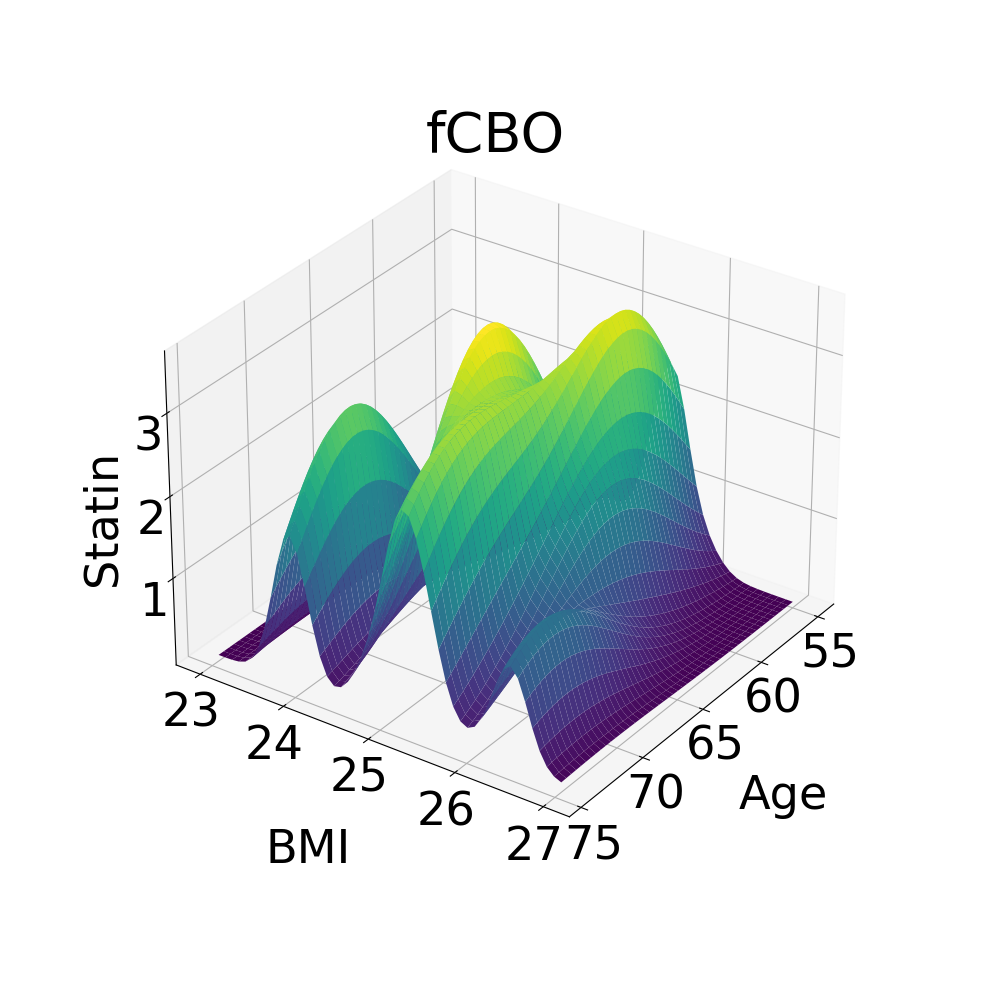}
    \hspace{-0.7cm}
    \includegraphics[width=0.23\textwidth]{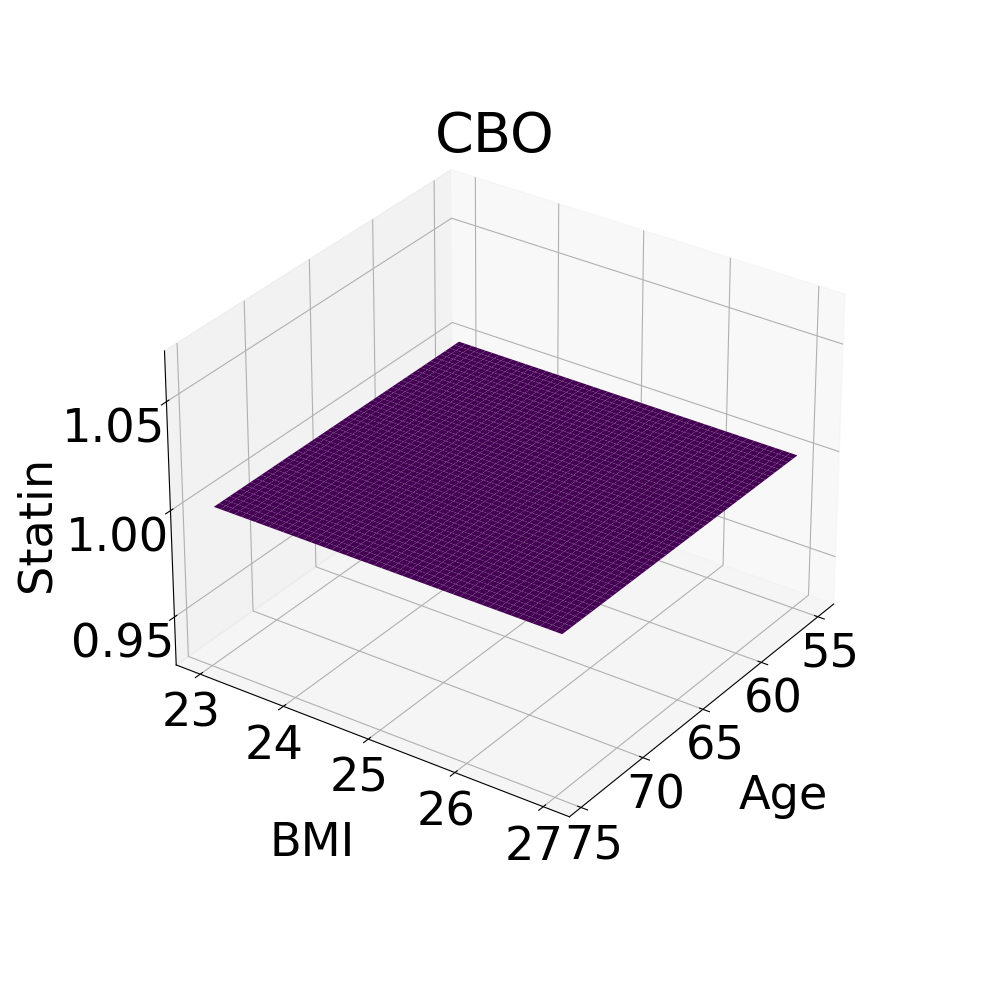}
    \includegraphics[width=3.3cm, height=3.3cm, keepaspectratio]{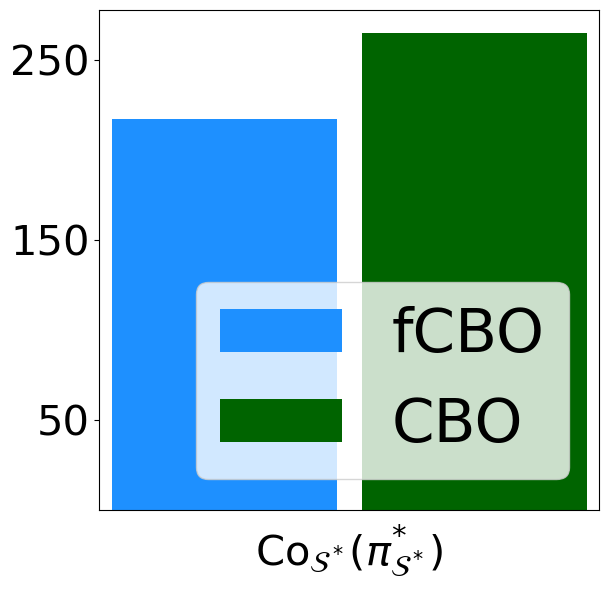}
    \caption{\health experiments with $\texttt{GridSize}=8$. \emph{Left}: Average convergence of \fcbo to the \fcgo optimum (solid red line); of \cbo, \bo, and \mcbo-\textsc{h} to the \cgo optimum (dotted red line); and of \fbo to the \fgo optimum (dashed red line) (across 20 different initializations of $\datai$ for \fcbo, \cbo, \bo, \fbo, and across 20 seeds for \mcbo-\textsc{h} -- shaded areas give $\pm$ standard deviation).
    \emph{Middle}: $\pi^*_{\Statin|\Age, \bmi}$ found by \fcbo (left) and $\pi^*_{\Statin|\emptyset}$ found by \cbo (right) across levels of \Age~and \bmi.
    \emph{Right}: Cost associated to the optimal \mps and associated optimal \dmp found by \fcbo and \cbo.}
    \label{fig:health_grid8_full_results}
\end{figure*}
Similar results are observed with $\texttt{GridSize} = 8$ (\figref{fig:health_grid8_full_results}). \fcbo achieves the smallest target effect (\figref{fig:health_grid8_full_results}, left), and incurs a lower cost compared to \cbo (\figref{fig:health_grid8_full_results}, right). In this setting \fcbo converges to $\Sset^\st =\{\langerangle{\Aspirin}{\emptyset}, \langerangle{\Statin}{(\text{Age}, \bmi)}\}$ with $\pi^*_{\Sset^*} = \{0.1,\pi^*_{\Statin|\text{Age}, \bmi}\}$. Due to the more complex  $\pi^*_{\Statin|\text{Age} \bmi}$ (\figref{fig:health_grid8_full_results}(middle, left)), which allocates the highest \Statin~dosages to mid-range value of \Age~and \bmi, the investigator can avoid intervening on \ci thus lowering the overall cost of the intervention 
while still achieving an overall smaller target effect. 

\section{Conclusion}\label{sec:conclusion}
We proposed the \fcbo method for finding policies made of hard and functional interventions that optimize a target effect. We introduced graphical criteria that establish when functional interventions could be necessary to achieve optimal target effects and when hard interventions are sufficient. Furthermore, we showed that optimizing a target effect by considering functional interventions allows the investigator to identify policies that are also optimal w.r.t. conditional target effects. We demonstrated the benefit of the proposed approach on a synthetic and on a real-world causal graph. Future work will explore the use of gradient-based optimization methods for the acquisition functional, as well as the development of more flexible kernel construction for the \gptext functionals (see Appendix \ref{sec:app:alternate_kernels}). 
These extensions would enable the identification of more flexible functional interventions while speeding up the convergence of the algorithm.

\begin{acknowledgements} 
The authors would like to thank Michalis Titsias, Alan Malek, and Eleni Sgouritsa for valuable discussions.
\end{acknowledgements}

\bibliography{bib}

\begin{thebibliography}{30}
\providecommand{\natexlab}[1]{#1}
\providecommand{\url}[1]{\texttt{#1}}
\expandafter\ifx\csname urlstyle\endcsname\relax
  \providecommand{\doi}[1]{doi: #1}\else
  \providecommand{\doi}{doi: \begingroup \urlstyle{rm}\Url}\fi

\bibitem[Aglietti et~al.(2020)Aglietti, Lu, Paleyes, and
  Gonz{\'{a}}lez]{aglietti2020causal}
Virginia Aglietti, Xiaoyu Lu, Andrei Paleyes, and Javier Gonz{\'{a}}lez.
\newblock Causal {B}ayesian optimization.
\newblock In \emph{International Conference on Artificial Intelligence and
  Statistics}, pages 3155--3164, 2020.

\bibitem[Aglietti et~al.(2021)Aglietti, Dhir, Gonz{\'a}lez, and
  Damoulas]{aglietti2021dynamic}
Virginia Aglietti, Neil Dhir, Javier Gonz{\'a}lez, and Theodoros Damoulas.
\newblock Dynamic causal {B}ayesian optimization.
\newblock In \emph{Advances in Neural Information Processing Systems}, pages
  10549--10560, 2021.

\bibitem[Aronszajn(1950)]{aronszajn1950theory}
Nachman Aronszajn.
\newblock Theory of reproducing kernels.
\newblock \emph{Transactions of the American Mathematical Society}, 68\penalty0
  (3):\penalty0 337--404, 1950.

\bibitem[Arsenyan et~al.(2023)Arsenyan, Grosnit, and
  Bou-Ammar]{arsenyan2023contextual}
Vahan Arsenyan, Antoine Grosnit, and Haitham Bou-Ammar.
\newblock Contextual causal {B}ayesian optimisation.
\newblock \emph{arXiv preprint arXiv:2301.12412}, 2023.

\bibitem[Correa and Bareinboim(2020{\natexlab{a}})]{correa2020calculus}
Juan Correa and Elias Bareinboim.
\newblock A calculus for stochastic interventions: Causal effect identification
  and surrogate experiments.
\newblock In \emph{AAAI Conference on Artificial Intelligence}, pages
  10093--10100, 2020{\natexlab{a}}.

\bibitem[Correa and Bareinboim(2020{\natexlab{b}})]{correa2020general}
Juan Correa and Elias Bareinboim.
\newblock General transportability of soft interventions: Completeness results.
\newblock In \emph{Advances in Neural Information Processing Systems}, pages
  10902--10912, 2020{\natexlab{b}}.

\bibitem[De~Kroon et~al.(2022)De~Kroon, Mooij, and Belgrave]{de2022causal}
Arnoud De~Kroon, Joris Mooij, and Danielle Belgrave.
\newblock Causal bandits without prior knowledge using separating sets.
\newblock In \emph{Conference on Causal Learning and Reasoning}, pages
  407--427, 2022.

\bibitem[Feng et~al.(2020)Feng, Letham, Mao, and Bakshy]{feng2020high}
Qing Feng, Ben Letham, Hongzi Mao, and Eytan Bakshy.
\newblock High-dimensional contextual policy search with unknown context
  rewards using {B}ayesian optimization.
\newblock In \emph{Advances in Neural Information Processing Systems}, pages
  22032--22044, 2020.

\bibitem[Ferro et~al.(2015)Ferro, Pina, Severo, Dias, Botelho, and
  Lunet]{ferro2015use}
Ana Ferro, Francisco Pina, Milton Severo, Pedro Dias, Francisco Botelho, and
  Nuno Lunet.
\newblock Use of statins and serum levels of prostate specific antigen.
\newblock \emph{Acta Urol{\'o}gica Portuguesa}, 32\penalty0 (2):\penalty0
  71--77, 2015.

\bibitem[Gasse et~al.(2021)Gasse, Grasset, Gaudron, and
  Oudeyer]{gasse2021causal}
Maxime Gasse, Damien Grasset, Guillaume Gaudron, and Pierre-Yves Oudeyer.
\newblock Causal reinforcement learning using observational and interventional
  data.
\newblock \emph{arXiv preprint arXiv:2106.14421}, 2021.

\bibitem[Koller and Friedman(2009)]{kollerl2009probabilistic}
Daphne Koller and Nir Friedman.
\newblock \emph{Probabilistic Graphical Models: Principles and Techniques}.
\newblock MIT Press, 2009.

\bibitem[Krause and Ong(2011)]{krause2011contextual}
Andreas Krause and Cheng Ong.
\newblock Contextual {G}aussian process bandit optimization.
\newblock In \emph{Advances in Neural Information Processing Systems}, 2011.

\bibitem[Lattimore et~al.(2016)Lattimore, Lattimore, and
  Reid]{lattimore2016causal}
Finnian Lattimore, Tor Lattimore, and Mark~D Reid.
\newblock Causal bandits: Learning good interventions via causal inference.
\newblock In \emph{Advances in Neural Information Processing Systems}, 2016.

\bibitem[Lee and Bareinboim(2018)]{lee2018structural}
Sanghack Lee and Elias Bareinboim.
\newblock Structural causal bandits: Where to intervene?
\newblock In \emph{Advances in Neural Information Processing Systems}, 2018.

\bibitem[Lee and Bareinboim(2019)]{lee2019structural}
Sanghack Lee and Elias Bareinboim.
\newblock Structural causal bandits with non-manipulable variables.
\newblock In \emph{AAAI Conference on Artificial Intelligence}, pages
  4164--4172, 2019.

\bibitem[Lee and Bareinboim(2020)]{lee2020characterizing}
Sanghack Lee and Elias Bareinboim.
\newblock Characterizing optimal mixed policies: Where to intervene and what to
  observe.
\newblock In \emph{Advances in Neural Information Processing Systems}, pages
  8565--8576, 2020.

\bibitem[Lu et~al.(2018)Lu, Sch{\"o}lkopf, and
  Hern{\'a}ndez-Lobato]{lu2018deconfounding}
Chaochao Lu, Bernhard Sch{\"o}lkopf, and Jos{\'e}~Miguel Hern{\'a}ndez-Lobato.
\newblock Deconfounding reinforcement learning in observational settings.
\newblock \emph{arXiv preprint arXiv:1812.10576}, 2018.

\bibitem[Lu et~al.(2020)Lu, Meisami, Tewari, and Yan]{lu2020regret}
Yangyi Lu, Amirhossein Meisami, Ambuj Tewari, and William Yan.
\newblock Regret analysis of bandit problems with causal background knowledge.
\newblock In \emph{Conference on Uncertainty in Artificial Intelligence}, pages
  141--150, 2020.

\bibitem[Nair et~al.(2021)Nair, Patil, and Sinha]{nair2021budgeted}
Vineet Nair, Vishakha Patil, and Gaurav Sinha.
\newblock Budgeted and non-budgeted causal bandits.
\newblock In \emph{International Conference on Artificial Intelligence and
  Statistics}, pages 2017--2025, 2021.

\bibitem[Pearl(2000)]{pearl2000causality}
Judea Pearl.
\newblock \emph{Causality: Models, Reasoning and Inference}.
\newblock Springer, 2000.

\bibitem[Rawla(2019)]{rawla2019epidemiology}
Prashanth Rawla.
\newblock Epidemiology of prostate cancer.
\newblock \emph{World Journal of Oncology}, 10\penalty0 (2):\penalty0 63, 2019.

\bibitem[Shahriari et~al.(2015)Shahriari, Swersky, Wang, Adams, and
  De~Freitas]{shahriari2015taking}
Bobak Shahriari, Kevin Swersky, Ziyu Wang, Ryan~P Adams, and Nando De~Freitas.
\newblock Taking the human out of the loop: A review of {B}ayesian
  optimization.
\newblock \emph{Proceedings of the IEEE}, 104\penalty0 (1):\penalty0 148--175,
  2015.

\bibitem[Shilton et~al.(2020)Shilton, Gupta, Rana, and
  Venkatesh]{shilton2020sequential}
Alistair Shilton, Sunil Gupta, Santu Rana, and Svetha Venkatesh.
\newblock Sequential subspace search for functional {B}ayesian optimization
  incorporating experimenter intuition.
\newblock \emph{arXiv preprint arXiv:2009.03543}, 2020.

\bibitem[Sussex et~al.(2023)Sussex, Makarova, and Krause]{sussex2022model}
Scott Sussex, Anastasiia Makarova, and Andreas Krause.
\newblock Model-based causal {B}ayesian optimization.
\newblock In \emph{International Conference on Learning Representations}, 2023.

\bibitem[Vellanki et~al.(2019)Vellanki, Rana, Gupta, Rubin~de Celis~Leal,
  Sutti, Height, and Venkatesh]{vellanki2019bayesian}
Pratibha Vellanki, Santu Rana, Sunil Gupta, David Rubin~de Celis~Leal,
  Alessandra Sutti, Murray Height, and Svetha Venkatesh.
\newblock Bayesian functional optimisation with shape prior.
\newblock In \emph{{AAAI} Conference on Artificial Intelligence}, pages
  1617--1624, 2019.

\bibitem[Vien et~al.(2018)Vien, Zimmermann, and Toussaint]{vien2018bayesian}
Ngo~Anh Vien, Heiko Zimmermann, and Marc Toussaint.
\newblock Bayesian functional optimization.
\newblock In \emph{AAAI Conference on Artificial Intelligence}, pages
  4171--4178, 2018.

\bibitem[Williams and Rasmussen(2006)]{williams2006gaussian}
Christopher~KI Williams and Carl~Edward Rasmussen.
\newblock \emph{Gaussian Processes for Machine Learning}.
\newblock MIT Press, 2006.

\bibitem[Zhang(2020)]{zhang2020designing}
Junzhe Zhang.
\newblock Designing optimal dynamic treatment regimes: A causal reinforcement
  learning approach.
\newblock In \emph{International Conference on Machine Learning}, pages
  11012--11022, 2020.

\bibitem[Zhang and Bareinboim(2019)]{zhang2019near}
Junzhe Zhang and Elias Bareinboim.
\newblock Near-optimal reinforcement learning in dynamic treatment regimes.
\newblock In \emph{Advances in Neural Information Processing Systems}, 2019.

\bibitem[Zhang and Bareinboim(2022)]{zhang2022online}
Junzhe Zhang and Elias Bareinboim.
\newblock Online reinforcement learning for mixed policy scopes.
\newblock In \emph{Advances in Neural Information Processing Systems}, 2022.

\end{thebibliography}

\appendix
\onecolumn

\section{Proofs}\label{sec:app:proofs}
\begin{namedthm*}{Proposition \ref{prop:hard_suboptimality}}
Let $\graph$ be a causal graph such that (i) $\exists C\in \pa_{\graph}(Y)$ with $C\notin \I$; or (ii) $\exists C\in \text{sp}_{\graph}(Y)$. If $\exists X \in \an_{\graph}(Y) \cap \I$ such that $\{\langle X, C \rangle\}$ is an \mps, then there exists at least one \scm compatible with $\graph$ for which $\min_{\Sset \in \Sigma_{\hardsubscript}, \piS \in \PiS} \mu^Y_{\piS}>\min_{\Sset \in \Sigma, \piS \in \PiS} \mu^Y_{\piS}$.
\end{namedthm*}

\begin{proof}
\emph{Case (i)}: 
Assume that there exists $C\in \pa_{\graph}(Y)$ with $C\notin \I$ and $X \in \an_{\graph}(Y) \cap \I$ such that $\{\langle X, C \rangle\}$ is an \mps. As $X \in \an_{\graph}(Y)$, there exists a directed path from $X$ to $Y$, say $X \rightarrow X_i \rightarrow X_{i-1} \rightarrow \cdots \rightarrow X_1 \rightarrow Y$ without loss of generality. Let ${\cal M} = \langle \V, \U, \calF, p(\U) \rangle$ be an \scm such that
\begin{align*}
&C = U_{C}, \, U_{C} \sim \mathcal N(0,1), \\
&X_i = X, \, X_{i-1} = X_i,\, \dots,\, X_1 = X_2, \\
&Y = X_1 C U_Y, \, U_Y \sim \mathcal N(1,1).
\end{align*}    
${\cal M}$ is compatible with $\graph$. 
In this \scm, any \dmp 
$\pi_\Sset$ with $\Sset \in \Sigma_{\hardsubscript}$ would give  $\mu_{\pi_\Sset}^Y=\mathbb{E}_{\pi_\Sset}[Y]=0$.
In contrast, a \dmp $\pi_\Sset$ including the functional intervention $\pi_{X|C}(C)= -1/C$ would result in $Y = - U_Y$ and therefore  $\mu_{\pi_\Sset}^Y = -1$, giving $\min_{\Sset \in \Sigma_{\hardsubscript}, \piS \in \PiS} \mu^Y_{\piS}=0>-1\geq \min_{\Sset \in \Sigma, \piS \in \PiS} \mu^Y_{\piS}$.

\emph{Case (ii)}: 
Assume that there exists $C\in \text{sp}_{\graph}(Y)$ and $X \in \an_{\graph}(Y) \cap \I$ such that $\{\langle X, C \rangle\}$ is an \mps. As $X \in \an_{\graph}(Y)$, there exists a directed path from $X$ to $Y$, say $X \rightarrow X_{i} \rightarrow X_{i-1} \rightarrow \cdots \rightarrow X_1 \rightarrow Y$ without loss of generality. Let ${\cal M} = \langle \V, \U, \calF, p(\U) \rangle$ be an \scm such that
\begin{align*}
&C = U_{CY},\, U_{CY} \sim \mathcal N(0,1), \\
&X_i = X, \, X_{i-1} = X_i,\, \dots,\, X_1 = X_2, \\
&Y = X_1 U_{CY} U_Y, \, U_Y \sim \mathcal N(1,1).
\end{align*}    
${\cal M}$ is compatible with $\graph$. In this \scm, any \dmp $\pi_\Sset$ with $\Sset \in \Sigma_{\hardsubscript}$ would give
$\mu_{\pi_\Sset}^Y=\mathbb{E}_{\pi_\Sset}[Y]=0$. In contrast, a \dmp $\pi_\Sset$ containing the functional intervention $\pi_{X|C}(C)= -1/C$, would result in $Y = - U_Y$ and therefore $\mu_{\pi_\Sset}^Y = -1$, giving $\min_{\Sset \in \Sigma_{\hardsubscript}, \piS \in \PiS} \mu^Y_{\piS}=0>-1\geq \min_{\Sset \in \Sigma, \piS \in \PiS} \mu^Y_{\piS}$. 
\end{proof}

In the following proposition we use the notation $\graph_{\underline{\X}}$ to indicate the modification of $\graph$ obtained by removing the outgoing edges from $\X$.

\begin{namedthm*}{Proposition \ref{prop:hard_optimality}}
In a casual graph $\graph$, if $\pa_{\graph}(Y)\subseteq \I$ and $\text{sp}_{\graph}(Y)=\emptyset$ there exists a \dmp compatible with \mps $\Sset=\{\langerangle{X}{\emptyset}: X \in \pa_{\graph}(Y)\}$ that solves the \fcgo problem.
\end{namedthm*}

\begin{proof}
Consider \mps $\Sset\in\Sigma$ for $\graph$ and \dmp $\pi_{\mathcal S}$ compatible with $\Sset$. Let $\Z = \pa_{\graph}(Y) \backslash ((\X_{\Sset}\cup \C_{\Sset}) \cap \pa_{\graph}(Y))$. As $\pa_{\graph}(Y)\subseteq \I$, 
we can define the \mps $\Sset_\pa = \{\langerangle{X}{\emptyset}: \forall X \in \pa_{\graph}(Y)\}$. Denote by $p_{\pi^*_{\Sset_{\pa}}}(Y)$ the distribution of $Y$ induced by an optimal \dmp $\pi^*_{\Sset_{\pa}}$ compatible with $\Sset_{\text{pa}}$, \ie such that 
$\int_{\range_Y} Y p_{\pi^\st_{\Sset_\pa}}(Y)dY\leq \int_{\range_Y} Y p_{\pi_{\Sset_\text{pa}}}(Y)dY$, for every \dmp $\pi_{\Sset_\text{pa}}$ compatible with $\Sset_{\pa}$, and let $\range = \range_Y \times \range_{\X_{\Sset}\cup \C_{\Sset}}\times \range_{\Z}$. Exploiting the rules of do-calculus \citep{pearl2000causality} and $\sigma$-calculus \citep{correa2020calculus} we obtain
\begin{align*}
\mu_{\pi_{\Sset}}^Y 
&= \int_{\range} Y p_{\pi_{\Sset}}(Y \cond \X_{\Sset}\cup \C_{\Sset} \cup \Z) \underbrace{p_{\pi_{\Sset}}(\X_{\Sset}\cup \C_{\Sset} \cup \Z)d\X_{\Sset}\cup\C_{\Sset} d\Z dY }_{{\cal A}}\\
&= \int_{\range} Y p_{\pi_{\Sset}}(Y \cond \pa_{\graph}(Y)) {\cal A}  \hskip2.8cm (\text{rule 1 }\sigma\text{-calculus}) \,\, \begin{small}Y \indep_{\graph_{\Sset}} (\X_{\Sset}\cup \C_{\Sset} \cup \Z)\backslash \pa_{\graph}(Y) \cond \pa_{\graph}(Y) \end{small}\\
&= \int_{\range} Y p(Y \cond \pa_{\graph}(Y)) {\cal A} \hskip3.2cm (\text{rule 2 }\sigma\text{-calculus}) \,\, \begin{small}Y \indep_{\graph_{\Sset, \underline{\X_{\Sset}}}, \graph_{ \underline{\X_{\Sset}}}} \X_{\Sset} \cond (\pa_{\graph}(Y)\backslash(\pa_{\graph}(Y) \cap \X_{\Sset})) \end{small} \\
&= \int_{\range} Y p(Y \cond \doi(\pa_{\graph}(Y))) {\cal A} \hskip2.6cm (\text{rule 2 } \text{do-calculus})\,\, \begin{small} Y \indep_{\graph_{\underline{\pa_{\graph}(Y)}}} \pa_{\graph}(Y)  \end{small} 
\\
&= \int_{\range} Y p_{\pi_{\Sset_{\pa}}}(Y) {\cal A} 
\geq \int_{\range} Y p_{\pi^\st_{\Sset_{\pa}}}(Y) {\cal A} 
= \mu^Y_{\pi^\st_{\Sset_{\text{pa}}}},
\end{align*}
where $\indep_{\graph_{\Sset, \underline{\X_{\Sset}}}, \graph_{ \underline{\X_{\Sset}}}}$ denotes d-separation in both $\graph_{\Sset, \underline{\X_{\Sset}}}$ and $\graph_{\underline{\X_{\Sset}}}$.
\end{proof}

\begin{namedthm*}{Proposition~\ref{prop:soft_opt}}
If  $\Sset^\st, \pi^{\st}_{\Sset^\st}=\argmin_{\Sset \in \Sigma, \piS\in\PiS}\mu^Y_{\piS}$, 
then $\Sset^\st, \pi^{\st}_{\Sset^\st}=\argmin_{\Sset \in \Sigma^{\C}, \piS\in\PiS}\mu^Y_{\piS,\C=\c}$ $\forall \C \subset \V \backslash Y$ 
such that $\C \cap \text{de}_{\graph}(\I)=\emptyset$ and $\forall\c \in \range_{\C}$ with $\Sigma^{\C} = \{\Sset \in \Sigma: \X_{\Sset} = \X_{\Sset^\st} \text{ and } \{\langerangle{X}{\C_X^{\Sset^\st} \cup \C_X^{\Sset} \cup \C}:X \in \X_{\Sset^\st}\} \text{ is an } \mps\}$.
\end{namedthm*}

\begin{proof}
Assume, by contradiction, that $(\Sset^\st, \pi^{\st}_{\Sset^\st})$, with $\pi^{\st}_{\Sset^\st}=\left\{ \pi_{X|\C_X^{\Sset^\st}}^{\Sset^\st} \right\}_{X\in \C_X^{\Sset^\st}}$, is a solution to the \fcgo problem but there exist $\C \subset \V \backslash Y$ and a value $\c \in \range_\C$ such that the tuple $(\Sset^1, \pi_{\Sset^1})$ with $\Sset^1 \in \Sigma^\C$ and $\pi_{\Sset^1}=\left\{ \pi_{X|\C_X^{\Sset^1}}^{\Sset^1} \right\}_{X\in \C_X^{\Sset^1}}\in \PiS$ satisfies $\mu^Y_{\pi_{\Sset^1}, \C=\c} < \mu^Y_{\pi^{\st}_{\Sset^\st},\C=\c}$. As $\Sset^1 \in \Sigma^\C$, we can construct \mps $\Sset^2 = \{\langerangle{X}{\C_X^{\Sset^\st} \cup \C_X^{\Sset^1} \cup \C}: X \in \X_{\Sset^\st}\}$ and the compatible  $\pi_{\Sset^2}=\left\{\pi^{\Sset^2}_{X|\C_X^{\Sset^\st} \cup \C_X^{\Sset^1} \cup \C}\right\}_{X \in \X_{\Sset^\st}}$ with 
\begin{align*}
\pi^{\Sset^2}_{X|\C_X^{\Sset^\st} \cup \C_X^{\Sset^1} \cup \C}=
\begin{cases}
& \pi_{X|\C_X^{\Sset^1}}^{\Sset^1} \text{ if } \C \in [\c-\delta,\c+\delta] \\
& \pi_{X|\C_X^{\Sset^\st}}^{\Sset^\st}  \text{ otherwise},
\end{cases}
\end{align*} 
for a small enough $\delta>0$. As $\C \cap \de_{\graph}(\I) =\emptyset$, variables in $\C$ are not affected by interventions on variables in $\X_{\Sset^\st}$, and therefore $p_{\pi^\st_{\Sset^\st}}(\C)= p_{\pi_{\Sset^1}}(\C)=p(\C)$. Thus we obtain:
\begin{align*}
    \mu^Y_{\pi_{\Sset^2}} &= \int_{\range_\C} \mu^Y_{\pi_{\Sset^2}, \C=\c'} \; p_{\pi_{\Sset^2}}(\C=\c')d\c'\\
    &= \int_{[\c-\delta,\c+\delta]} \mu^Y_{\pi_{\Sset^2}, \C=\c'}\; p_{\pi_{\Sset^2}}(\C=\c')d\c' +  \int_{\range_\C\backslash [\c-\delta,\c+\delta]} \mu^Y_{\pi_{\Sset^2}, \C=\c'}\; p_{\pi_{\Sset^2}}(\C=\c')d\c'\\
    &= \int_{[\c-\delta,\c+\delta]} \mu^Y_{\pi_{\Sset^1}, \C=\c'}\; p_{\pi_{\Sset^1}}(\C=\c')d\c' + \int_{\range_\C\backslash [\c-\delta,\c+\delta]} \mu^Y_{\pi^{\st}_{\Sset^\st}, \C=\c'} \; p_{\pi^{\st}_{\Sset^\st}}(\C=\c')d\c'\\
    &< \int_{[\c-\delta,\c+\delta]} \mu^Y_{\pi^{\st}_{\Sset^\st}, \C=\c'}\; p_{\pi^{\st}_{\Sset^\st}}(\C=\c')d\c' + \int_{\range_\C\backslash [\c-\delta,\c+\delta]} \mu^Y_{\pi^{\st}_{\Sset^\st}, \C=\c'} \; p_{\pi^{\st}_{\Sset^\st}}(\C=\c')d\c'\\
    &=\mu^Y_{\pi^{\st}_{\Sset^\st}},
\end{align*}
with contradicts the assumption that $(\Sset^\st, \pi^{\st}_{\Sset^\st})$ is a solution to the \fcgo problem.
\end{proof}

\section{Alternative kernel construction}\label{sec:app:alternate_kernels}
The kernel function $\kappa_{\Sset}^\kappapar$ introduced in \secref{sec:gpsurrogate} sets the covariance between the elements in the vector $\pi_{\softsubscript}$ associated to a \dmp $\pi_{\Sset}$ to 0, thus restricting the type of functions that can be selected during optimization\footnote{Notice that, for hard interventions, this corresponds to limiting the range of values that can be set when intervening.}.

\begin{wrapfigure}[4]{r}{0.18\textwidth}
\vskip-0.5cm
\scalebox{0.9}{
\begin{tikzpicture}[dgraph]
\node[dot] (c1) [fill=darkGreen!70,label=north:$C_1$] at (-0.8, 2) {};
\node[dot] (c2) [fill=darkGreen!70,label=north:$C_2$] at (0.8, 2) {};
\node[dotdot node] (x) [fill=brightBlue!70,label=north:$X$] at (0, 1.4) {};
\node[dotdot node] (z)[fill=brightBlue!70,label=north:$Z$] at (1.6,1.4) {};
\node[dot] (y) [fill=red!70,label=north:$Y$] at (0.8,0.8) {};
\draw[line width=0.6pt, brightBlue, \arr](c1)--(x);
\draw[line width=0.6pt, brightBlue, \arr](c2)--(x);
\draw[line width=0.6pt, brightBlue, \arr](c2)--(z);
\draw[line width=0.6pt, \arr](x)--(y);
\draw[line width=0.6pt, \arr](z)--(y);
\end{tikzpicture}}
\end{wrapfigure}

For instance, consider the graph on the right with $\Sset = \{\langerangle{X}{(C_1, C_2)}, \langerangle{Z}{C_2}\}$ and $\piS = \{\pi_{X|\{C_1, C_2\}}, \pi_{Z|C_2}\}$. The proposed kernel function would set $\text{Cov}(\pi_{X|\{C_1, C_2\}}, \pi_{Z|C_2}) = 0$. While a study of the effect of choosing different covariance structures on the optimal target effect goes beyond the scope of this paper, in this section we provide alternative kernel constructions that relax this constraint.

Given a \dmp $\pi_\Sset$, one can define the correlation between elements in $\pi_{\softsubscript}$ by introducing a $|\C_\Sset|$-dimensional vector $\boldomega$ of parameters for each function $\pi_{X|\C_X}$ in $\pi_{\softsubscript}$ such that the $j$-th term $\omega_j=1$ if the $j$-th term in $\C_\Sset$ is in $\C_{X}$ and $\omega_j=0$ otherwise. For instance, for $\piS = \{\pi_{X|\{C_1, C_2\}}, \pi_{Z|C_2}\}=\pi_{\functsubscript}$, we have $\omega_1 = \omega_2 = 1$ for $\pi_{X|\{C_1, C_2\}}$ as both variables in $\C_{\Sset} = \{C_1, C_2\}$ are in $\C_X$, while $\omega_1 = 0$ and $\omega_2 = 1$ for $\pi_{Z|C_2}$ as only $C_2$ is in $\C_Z$. 

We can then redefine $\kappa_{\Sset}^\kappapar$ to be an \rbf kernel on an input space given by product between the the context variables and the $\boldomega$ parameters. Denote by $\boldomega^i, \boldomega^j$ two possible values for the $\boldomega$ vector, for instance we could have $\boldomega^i=[1, 1]^\top$ and $\boldomega^j=[0, 1]^\top$ in the example above; and by $\c^i = [c_1^i, \dots, c_{|\C_{\Sset}|}^i]^\top$ and $\c^j = [c_1^j, \dots, c_{|\C_{\Sset}|}^j]^\top$ two vector of values for $\C_{\Sset}$. We can define $\kappa_{\Sset}^\kappapar : (\range_{\C_{\Sset}} \times \Omega) \times (\range_{\C_{\Sset}} \times \Omega) \to \mathbb{R}^{|\Sset_{\functsubscript}| \times |\Sset_{\functsubscript}|}$ where $\Omega$ is the space of values for each vector $\boldomega$ and $\kappa_{\Sset}^\kappapar((\c, \boldomega)^i, (\c, \boldomega)^j) = \kappa_{\Sset}^\kappapar((\c^i)^\top \boldomega^i, (\c^j)^\top \boldomega^j) = \gamma \exp(-0.5/l^2 \sum_{n=1}^{|\C_{\Sset}|}(c^i_n\omega^i_n - c^j_n\omega^j_n)^2)$ where $\kappapar = \{\gamma, l\}$. For the example above, we can write $\kappa_{\Sset}^\kappapar((\c^i)^\top \boldomega^i, (\c^j)^\top \boldomega^j) = \gamma \exp(-0.5/l^2 [(c_1^i\omega_1^i - c_1^j\omega_1^j)^2 + (c_2^i\omega_2^i - c_2^j\omega_2^j)^2])$. When $\gamma\neq0$, $\boldomega^i=[1, 1]^\top$ and $\boldomega^j=[0, 1]^\top$, this kernel would return a covariance between $\pi_{X| C_1, C_2}$ and $\pi_{Z| C_2}$ equal to $\kappa_{\Sset}^\kappapar((\c^i)^\top \boldomega^i, (\c^j)^\top \boldomega^j) = \gamma \exp(-0.5/l^2 [(c_1^i)^2 + (c_2^i - c_2^j)^2])$. The covariance would thus depend on the context values in the overlapping part of the context variables space and a correction term $(c_1^i)^2$. Instead of fixing the values in $\boldomega$ to either zero or one based on the graph structure, one could think about optimizing the values that are different from zero so as to achieve a higher flexibility in terms of allowed covariance while still imposing structure via the zero values.

As a more general kernel construction, given a \dmp $\Sset$, a vector of parameter values $\boldomega^i$ and a vector of context values $\c^i = [c_1^i, \dots, c^i_{|\C_{\Sset}|}]^\top$, one could define the augmented input vector $\c^i_\text{aug} = [(\c^i)^\top \boldomega^i, (\c^i) \boldomega^i, t]^\top$ (and similarly for two alternative vector of values $\c^j$ and $\boldomega^j$) given by the concatenation of two $|\C_\Sset|$-dimensional vector obtained by $(\c^i)^\top \boldomega^i$ and a task index $t$ that gives the index of the function in $\pi_{\Sset_{\functsubscript}}$, similarly to what was introduced in \secref{sec:gpsurrogate}. 

For an augmented vector of hyper-parameters $\kappapar = [\gamma, l, \tilde{\gamma}, \tilde{l}]$, one could then define the following kernel:
\begin{align}
    \kappa_{\Sset}^{\kappapar}(\c_\text{aug}^i, \c_\text{aug}^j) &= \mathbb{I}_{t = t'} \gamma^2 \exp\left(-\frac{0.5}{l^2} \sum_{n=1}^{|\C_{\Sset}|} (\c^i_{\text{aug},n} - \c^j_{\text{aug}, n})^2\right)
    + \mathbb{I}_{t\neq t'} \tilde{\gamma}^2 \exp\left(-\frac{0.5}{\tilde{l}^2} \sum_{n=|\C_{\Sset}|+1}^{2|\C_{\Sset}|} (\c^i_{\text{aug},n} - \c^j_{\text{aug}, n})^2\right) \nonumber\\
    &=\mathbb{I}_{t = t'} \gamma^2 \exp\left(-\frac{0.5}{l^2} \sum_{n=1}^{|\C_{\Sset}|} (c^i_n\omega_n - c^j_n\omega'_n)^2\right)
    + \mathbb{I}_{t\neq t'} \tilde{\gamma}^2 \exp\left(-\frac{0.5}{\tilde{l}^2} \sum_{n=|\C_{\Sset}|+1}^{2|\C_{\Sset}|} (c^i_n\omega_n - c^j_n\omega'_n)^2\right),
    \label{eq:complex_kernel}
\end{align}
where $c^i_n$ is the $n$-th term of the $\c^i$ vector (similarly for $\c^j$ and $\boldomega^i$), and  $\mathbb{I}_{t=t'}$ is an indicator function equal to one if $t=t'$ and zero otherwise. The first term in \eqref{eq:complex_kernel} represents an \acro{rbf} kernel capturing the covariance structure \emph{within} the $t$-th function in $\pi_{\functsubscript}$ while the second term is again an \acro{rbf} kernel that captures the covariance \emph{across} functions in $\pi_{\functsubscript}$. Differently from the kernel described above we now have two sets of hyper-parameters: $\gamma, l$ for the first $\rbf$ kernel and $\tilde{\gamma}, \tilde{l}$ for the second. This gives higher flexibility in terms of the functional interventions we can learn and thus the target effect values we can achieve. As in the previous kernel we can let the parameters in $\boldomega$, as well as in $\kappapar$, change to capture different level of correlations or set them equal to one and zero depending on the structure of the graph. In the latter case and for the example introduced above, we would have $\omega_1 = \omega_2 = 1$ for $\pi_{X|C_1, C_2}$ which would lead to a standard \rbf kernel for the first term in \eqref{eq:complex_kernel}. We could then set $\tilde{\gamma}=0$ to have a zero covariance across functions or finally vary $\omega_3$ and $\omega_4$ for both $\pi_{X|C_1, C_2}$ and $\pi_{Z|C_2}$ to allow for increasing level of correlation.

\section{Chain Experiments}\label{sec:app:echain}
For the \echain experiments we use the following \scm: 
\begin{align*} 
& X = U_X, \hskip0.1cm W = U_W, \hskip0.1cm Z = -0.5X + U_Z, \hskip0.1cm Y = -W -3ZX + U_Y, \hskip0.1cm \text{with } U_X, U_W, U_Z, U_Y \sim {\cal N}(0,1).
\end{align*}
We set the range for hard interventions on both $Z$ and $W$ to $[-1, 1]$. The set of non-redundant \mps{s} is  $\mismps = \{\{\langerangle{Z}{\emptyset}\}, \{\langerangle{W}{\emptyset}\}, \{\langerangle{Z}{\emptyset}, \langerangle{W}{\emptyset}\}, \{\langerangle{Z}{\{X\}}\}, \{\langerangle{Z}{\{X\}}, \langerangle{W}{\emptyset}\}\}$. 

We set $\texttt{GridSize} = 10$ and represent each functional intervention with $N_{\alpha}=N_{\beta}=10$ samples for the context variables. We sample the coefficients $\boldalpha_i$ (for $i=1,\dots, N_\alpha$) and $\boldbeta_j$ (for $j=1, \dots, N_\beta$) uniformly in the interval $[-0.27, 0.27]$, in order to keep the range of values obtained for the intervened variables following a functional intervention similar to the ranges set for the hard interventions. For each $\Sset \in \mismps$, we initialize the linear kernel $\kappa^{\kappapar}_\Sset$ with $\kappapar = 1$. Exploration is hard to achieve when the $\gptext$ models for $\Sset$ including functional interventions are initialized with \rbf $K^{\theta}_\Sset$ and hyper-parameters $\theta = (\ell, \sigma^2_f) = (1, 1)$. We thus perform hyper-parameters search exploring continuous values $\sigma^2_f \in [1, 10000]$ and $\ell \in [1, 30]$, which results in selecting $\sigma_f^2 = 7000$, and $\ell = 20$ for both \fcbo and \fbo.
For \cbo and \bo, which consider only hard interventions and thus do not suffer from exploration issues, we initialize $K^{\theta}_\Sset$ with $\theta = (1, 1)$. For \mcbo we use the default setting (Mat\'ern $5/2$ kernel), as it is not possible to tune the kernel and corresponding hyper-parameters. In order to run \mcbo with contextual interventions, we use the augmented \scm  with action variables $X = U_X$, $W= U_W + A_W$, $Z = -0.5X + U_Z + A_Z$, $Y = -W -3ZX + U_Y$. 
In this setting, the average \acro{cpu} execution time for a single \fcbo run is $\sim$ 6 minutes, while for a single \mcbo run is $\sim$ 14 minutes.

\section{Health Experiments}\label{sec:app:health}
For the \health experiments, we use the \scm from \cite{ferro2015use}:
\begin{equation*}
\begin{split}
    &\Age = U_{\Age}, \ci = U_{\ci},\bmr = 1500 + 10 \times U_{\bmr}, \\
    &\text{Height} = 175 + 10 \times U_{\text{Height}}, \\
    &\text{Weight} = \frac{\bmr + 6.8 \times \text{Age} - 5 \times  \Height}{13.7 + \ci \times 150/7716}, \\
    &\bmi = \text{Weight} / (\text{Height}/100)^2, \\
    &\Aspirin = \sigma(-8 + 0.1 \times \Age + 0.03 \times \bmi), \\
    &\Statin = \sigma(-13 + 0.1 \times \Age + 0.2 \times \bmi), \\
    &\psa = 6.8 + 0.04 \times \Age - 0.15 \times \bmi - 0.6 \times \Statin + 0.55 \times \Aspirin \\
    & \hskip0.9cm+\sigma(2.2 - 0.05 \times \Age + 0.01\times \bmi - 0.04 \times \Statin + 0.02 \times \Aspirin) + U_{\psa}, 
\end{split}
\end{equation*}
with $U_{\text{Age}} \sim \mathcal{U}(55, 75)$,  $U_{\text{\acro{ci}}} \sim \mathcal{U}(-100, 100)$, $U_{\text{\acro{bmr}}} \sim  t\mathcal{N}(-1, 2)$, $U_{\text{Height}}\sim t\mathcal{N}(-0.5, 0.5)$, $U_{\acro{psa}} \sim \mathcal{N}(0, 0.4)$, where $\mathcal{U}(\cdot, \cdot)$ denotes a uniform distribution, $t\mathcal{N}(a, b)$ a standard Gaussian distribution truncated between $a$ and $b$, and $\sigma(\cdot)$ the sigmoidal transformation defined as $\sigma(x) = \frac{1}{1 + \exp(-x)}$. 

We set the ranges for hard interventions on Aspirin, Statin, and CI to $[0.1, 1]$.
The set of non-redundant \mps{s} is $\mismps =$ \{\{$\langerangle{\Aspirin}{\emptyset}$\},
\{$\langerangle{\Statin}{\emptyset}$\}, \{$\langerangle{\ci}{\emptyset}$\},
\{$\langerangle{\Aspirin}{\emptyset}$, $\langerangle{\Statin}{\emptyset}$\}, \{$\langerangle{\Aspirin}{\emptyset}$, $\langerangle{\ci}{\emptyset}$\}, \{$\langerangle{\Statin}{\emptyset}$, $\langerangle{\ci}{\emptyset}$\}, \{$\langerangle{\Aspirin}{\emptyset}$, $\langerangle{\Statin}{\emptyset}$, $\langerangle{\ci}{\emptyset}$\}, \{$\langerangle{\Aspirin}{\{\Age, \bmi\}}$\}, \{$\langerangle{\Statin}{\{\Age, \bmi\}}$\}, \{$\langerangle{\Aspirin}{\{\Age,\bmi\}}$, $\langerangle{\Statin}{\{\Age, \bmi\}}$\}, \{$\langerangle{\Aspirin}{\{\Age, \bmi\}}$, $\langerangle{\Statin}{\emptyset}$\}, \{$\langerangle{\Aspirin}{\emptyset}$, 
$\langerangle{\Statin}{\{\Age, \bmi\}}$\},\{$\langerangle{\Aspirin}{\{\Age, \bmi\}}$, $\langerangle{\ci}{\emptyset}$\},\{$\langerangle{\Statin}{\{\Age, \bmi\}}$, $\langerangle{\ci}{\emptyset}$\}, \{$\langerangle{\Aspirin}{\{\Age, \bmi\}}$, $\langerangle{\Statin}{\{\Age, \bmi\}}$, $\langerangle{\ci}{\emptyset}$\}, \{$\langerangle{\Aspirin}{\emptyset}$, $\langerangle{\Statin}{\{\Age, \bmi\}}$, $\langerangle{\ci}{\emptyset}$\}, \{$\langerangle{\Aspirin}{\{\Age, \bmi\}}$, $\langerangle{\Statin}{\emptyset}$, $\langerangle{\ci}{\emptyset}$\}\}. 

We represent each functional intervention with $N_{\alpha}=N_{\beta}=10$ samples for the context variables. We sample the coefficients $\boldalpha_i$ (for $i=1,\dots, N_\alpha$) and $\boldbeta_j$ (for $j=1, \dots, N_\beta$) uniformly in the interval $[0, 3.3]$, in order to keep the total cost of functional interventions and hard interventions comparable. The \rbf kernels $K^{\theta}_\Sset$ and $\kappa_{\Sset}^{\kappapar}$ are initialized with $\theta = (1, 1)$ and $\kappapar = (1, 1)$ for each $\Sset \in \mismps$. In this setting, the average \acro{cpu} execution time for a single \fcbo run is $\sim$ 3 hours and 20 minutes, while for a single \mcbo run is $\sim$ 10 hours.
\end{document}